\documentclass{article} 
\usepackage{arxiv_preprint,times}
\arxivfinalcopy
\usepackage{natbib}
\usepackage[utf8]{inputenc} 
\usepackage[T1]{fontenc}    
\usepackage{hyperref}       
\usepackage{url}            
\usepackage{booktabs}       
\usepackage{amsfonts}       
\usepackage{nicefrac}       
\usepackage{microtype}      
\usepackage{graphicx}         
\usepackage{float} 
\usepackage{xspace}
\usepackage{subcaption}
\usepackage{mwe} 
\usepackage{amsmath}
\usepackage{mathtools}
\usepackage{amssymb}
\usepackage{amsthm}
\usepackage{enumitem}
\usepackage{wrapfig}
\usepackage{tikz}
\usetikzlibrary{positioning, shapes, patterns, arrows.meta}
\usepackage[linesnumbered,ruled,vlined]{algorithm2e}
\usepackage[english]{babel}

\newlist{compactitem}{itemize}{1}
\setlist[compactitem,1]{label=\textbullet, left=0pt, itemsep=1pt, topsep=1pt, parsep=0pt, partopsep=0pt}
\DontPrintSemicolon
\SetAlgoNoEnd
\SetKwInOut{KwIn}{Input}
\SetKwInOut{KwOut}{Output}

\providecommand{\vs}{vs. }
\providecommand{\ie}{\emph{i.e.,} }
\providecommand{\eg}{\emph{e.g.,} }

\providecommand{\etc}{\emph{etc.}}

\providecommand{\myparab}[1]{\noindent\textbf{#1}}

\newcommand{\sysname}{Straggl\textsc{AR}\xspace}
\providecommand{\alltoall}{\textsc{AllToAll}\xspace}
\providecommand{\allreduce}{\textsc{AllReduce}\xspace}
\providecommand{\allgather}{\textsc{AllGather}\xspace}

\providecommand{\reducescatter}{\textsc{ReduceScatter}\xspace}

\title{Efficient \allreduce with Stragglers}

\author{Arjun Devraj, Eric Ding, Abhishek Vijaya Kumar, Robert Kleinberg, Rachee Singh\\
Cornell University\\
\texttt{adevraj@cs.cornell.edu}}
\begin{document}
    \maketitle
    \begin{abstract}
Distributed machine learning workloads use data and tensor parallelism for training and inference, both of which rely on the \allreduce{} collective to synchronize gradients or activations. However, \allreduce{} algorithms are delayed by the slowest GPU to reach the synchronization barrier before the collective (\ie the straggler). To address this challenge, we propose \sysname{}: a parallel algorithm for \allreduce that accelerates distributed training and inference by exploiting natural variation in GPU execution times. \sysname{} implements a \reducescatter{} among the remaining GPUs during the straggler-induced delay, and then executes a novel collective algorithm to complete the \allreduce{} once the final GPU reaches the synchronization barrier. 
\sysname achieves a $2\times$ theoretical speedup over popular bandwidth-efficient algorithms for large GPU clusters, surpassing the lower bound for bandwidth-optimal synchronous \allreduce by leveraging the asymmetry in when GPUs reach the synchronization barrier. On an 8-GPU server, \sysname{} provides a 25\% speedup over state-of-the-art \allreduce{} algorithms.
\end{abstract}
    \begin{figure}[h]
    \centering
    \includegraphics[width=0.7\textwidth]{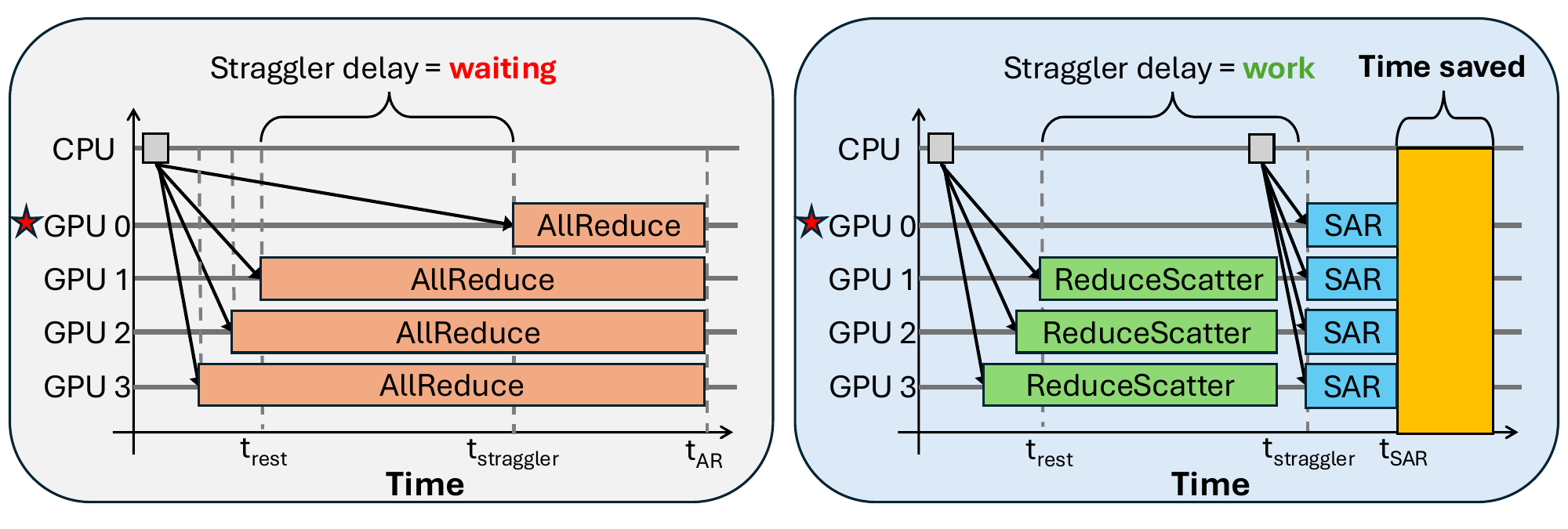}
    \caption{Straggler GPU (rank 0) causes all other GPUs to wait to begin the \allreduce operation (left). Our proposed straggler-aware \allreduce (right). Only communication kernels are shown.}
    \label{fig:straggler-diagram}
    \vspace{-1em}
\end{figure}

\section{Introduction}
\label{sec:intro}

Distributed training and inference rely on collective communication primitives to exchange model gradients and activations across multiple GPUs. In particular, the \allreduce primitive is used to average gradients across GPUs during data-parallel training and aggregate partial activations in tensor-parallel training and inference. \allreduce and other communication primitives are implemented in collective communication libraries (CCLs), like NVIDIA's NCCL, which provide the core communication infrastructure for distributed ML.

To enable efficient communication, CCLs implement \allreduce using optimized parallel algorithms rooted in decades of research in high-performance computing~\citep{mpi}. These algorithms are designed to minimize communication time across GPUs --- a key bottleneck in scaling modern ML workloads with massive data transfer sizes. To meet this demand, CCLs use bandwidth-optimal algorithms (\eg Ring, Recursive Halving/Doubling), which heavily parallelize communication to fully leverage available inter-GPU network bandwidth. 

\begin{figure}[t]
    \centering
     \begin{subfigure}[t]{0.44\linewidth}
        \centering
        \includegraphics[width=\textwidth]{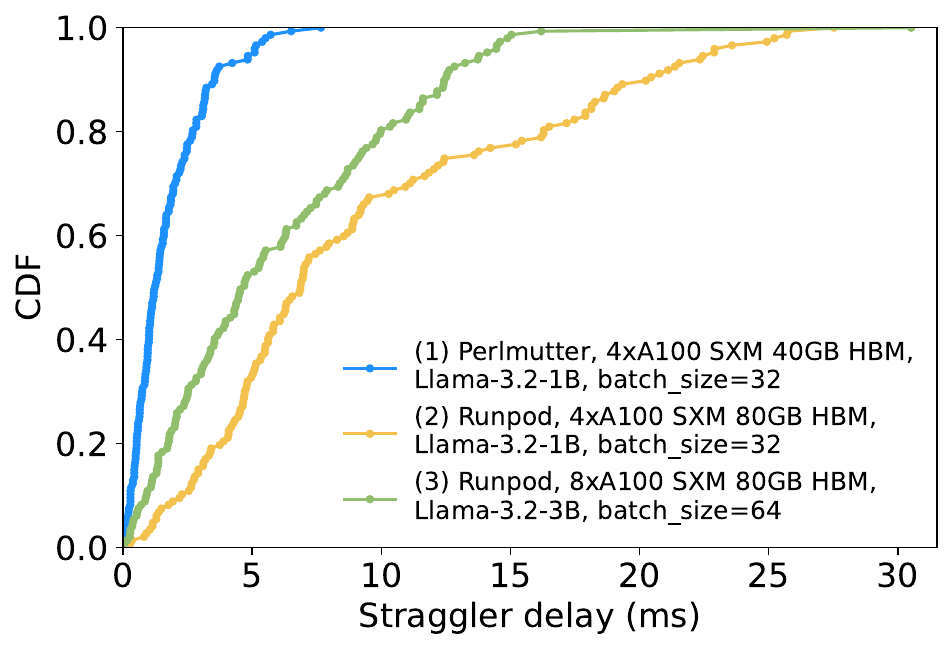}
        \caption{Straggler delays in Llama-3.2 fine-tuning jobs.}
        \label{fig:intro-motivation}
    \end{subfigure}
    \begin{subfigure}[t]{0.45\linewidth}
        \centering
        \includegraphics[width=\textwidth]{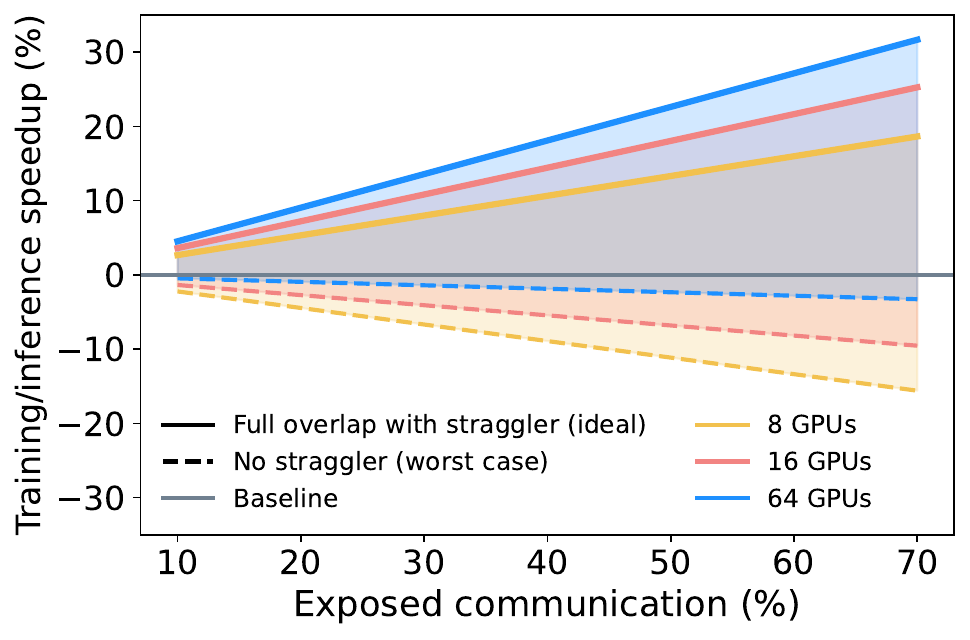}
        \caption{Range of simulated end-to-end speedups.}
        \label{fig:e2e-projections}
    \end{subfigure}
    \caption{\small{(a) CDFs of the straggler delay—the time between when the slowest and second-slowest ranks initiate the \allreduce{}—in Llama-3.2 (1B, 3B) fine-tuning jobs on the Perlmutter supercomputer and RunPod VMs (with 3 independent runs per job). (b) End-to-end speedups for ML workloads using \sysname{} compared to bandwidth-optimal algorithms (\eg Ring) for a 4 GB buffer (data-parallel buffer size of Llama-3.2-3B with local batch size of 4). Results are simulated with the $\alpha{-}\beta$ model using empirical $\alpha$ and $\beta$ values for the NVIDIA H100 DGX. Exposed communication is the percent of end-to-end time spent on \allreduce{}. As cluster size scales, \sysname{} shows large gains (+30\%) while worst-case performance is on par with the baseline (-3\%).}}
    \label{fig:intro-fig}
    \vspace{-2em}
\end{figure}

\myparab{Our key insight.} Today's collective algorithms are built on a fundamental assumption: all GPUs initiate the collective \emph{simultaneously}. However, in distributed ML, communication depends on preceding computation; the last GPU to finish the preceding computation, \ie the straggler, delays the entire \allreduce (Fig.~\ref{fig:straggler-diagram}, left). Our experiments show that this happens regularly even within a \emph{scale-up domain} --- the high-performance multi-GPU servers (\eg NVIDIA DGX) and racks (\eg NVIDIA GB200) that are the workhorses of distributed ML. In our experiments fine-tuning Llama-3.2 (1B, 3B) within DGX servers, we observe straggler delays of up to 30 milliseconds (Fig.~\ref{fig:intro-motivation}), prolonging the \allreduce execution time and causing other GPUs to stall. While stragglers in large datacenters are associated with network delays and hardware faults~\citep{wu2024falcon,lin2025understanding,warraich2025optireduce,jiang2024megascale}, we observe stragglers as an inherent challenge in distributed computation, where execution times naturally vary across GPUs. Existing mitigation strategies that approximate or drop the straggler's data can impact model convergence and do not generalize to \allreduce in tensor-parallel training/inference. Thus, there remains a pressing need for bandwidth-efficient \allreduce that is robust to stragglers while still preserving correctness.

\myparab{Our proposal.} Instead of treating stragglers as an anomaly, we propose designing algorithms that expect and exploit them. We design a novel \allreduce{} algorithm, \sysname{} (Fig.~\ref{fig:straggler-diagram}, right), that provably transmits up to $2\times$ fewer bytes than the known bandwidth-optimal lower bound by exploiting natural variation in GPU execution times. \sysname leverages the straggler delay to perform useful communication, eagerly executing a \reducescatter among the other GPUs. However, the resulting asymmetry between the data buffers of the straggler and non-straggler GPUs after the \reducescatter introduces new challenges in implementing highly parallel communication that can outperform known bandwidth-optimal algorithms. Thus, we design a new parallel communication algorithm that utilizes this asymmetry to perform faster \allreduce, achieving provably lower communication complexity in straggler settings while closely matching the complexity of bandwidth-optimal algorithms \emph{even without stragglers} (Fig.~\ref{fig:e2e-projections}). To our knowledge, our work is the first to show that the decades-old lower bound for bandwidth-optimal bulk-synchronous \allreduce{}~\citep{rabenseifner2004optimization,thakur2005optimization, patarasuk2009bandwidth,de2024swing} can be surpassed by leveraging natural variation in GPU compute times, while still maintaining the correctness guarantees of synchronous communication.

Our hardware experiments on 8-GPU servers show that \sysname achieves a 25\% speedup over bandwidth-optimal \allreduce algorithms (\eg Ring) in the presence of stragglers, and delivers end-to-end training speedups across multiple ML models. Simulations at larger scales show that these improvements grow with cluster size, reaching up to 2$\times$ speedups at scale. As shown in Fig.~\ref{fig:e2e-projections}, \sysname{'s} performance range shifts upwards with cluster size, yielding clear upside with variation in execution times and minimal downside otherwise. Finally, this work opens a new design dimension for collective algorithms: temporal asymmetry. For decades, we have pursued spatial optimizations like topology-aware routing~\citep{taccl} and spectral optimizations (\eg compression), but we have insisted on temporal symmetry (\ie all GPUs start the collective together). Breaking this assumption not only provides an algorithmic approach to mitigating straggler effects, but also presents an opportunity to fundamentally redesign the collective algorithms that underpin distributed ML. Our code is available at \url{https://github.com/arjundevraj/stragglar}.

\section{Background and Related Work}
\label{sec:background} 

\begin{wrapfigure}{r}{0.32\textwidth}
    \centering
    \includegraphics[width=0.32\textwidth, keepaspectratio]{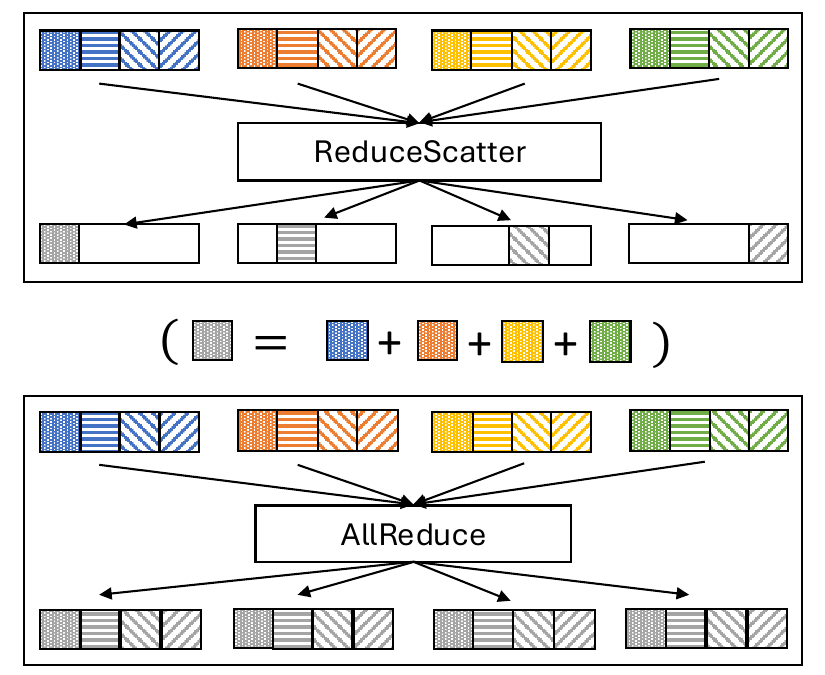}
    \caption{Collective operations.}
    \label{fig:collectives}
    \vspace{-15pt}
\end{wrapfigure}

In distributed ML, GPUs aggregate gradients and activations using the \allreduce collective communication primitive (Figure~\ref{fig:collectives}), which transmits and reduces data across the inter-GPU network. Data parallelism uses \allreduce to average gradients, while tensor parallelism invokes it many times per model pass to exchange activations~\citep{megatron}. Collective communication libraries (CCLs) like NCCL implement these primitives, choosing bandwidth-optimal algorithms (\eg Ring) for the large buffer sizes common in modern ML workloads~\citep{taccl}. CCLs adopt a bulk-synchronous model where all GPUs must synchronize before the collective, but performance degrades when a straggler---the slowest GPU---delays synchronization~\citep{warraich2025optireduce,dean2013tail,wang2024towards,gangidi2024rdma,li2014tales}. While stragglers result from inherent heterogeneity in GPU execution times, \emph{severe} stragglers can stem from hardware issues (thermal throttling, power supply) or runtime factors (network congestion, compute skew)~\citep{wu2024falcon,jiang2024megascale,grattafiori2024llama,xiong2024superbench}. Recent work has highlighted the acute impact of stragglers at datacenter scale~\citep{jiang2024megascale,lin2025understanding}, and our experiments show intrinsic 30 ms delays even within multi-GPU servers (Fig.~\ref{fig:intro-motivation}).

\myparab{Related Work.} Prior straggler mitigation strategies either identify and remove stragglers~\citep{jiang2024megascale,lin2025understanding}, wasting compute and only addressing severe cases, or approximate/drop the straggler's data~\citep{warraich2025optireduce,harlap2016addressing,karakus2017straggler,recht2011hogwild}, limiting applicability to data-parallel training and affecting convergence. Systems approaches~\citep{wu2024falcon,zhao2024adapcc} adapt workload placement or select among known algorithms, \eg AdapCC~\citep{zhao2024adapcc} uses known Tree algorithms that sacrifice bandwidth for latency. In contrast, \sysname is a novel bandwidth-efficient \allreduce algorithm that preserves exact reductions for both data and tensor parallelism and treats stragglers as intrinsic to distributed computation. While recent works synthesize new collective algorithms for heterogeneous networks~\citep{taccl,zhao2024adapcc,wang2020blink,won2024tacos,zhang2024rethinking}, in homogeneous scale-up domains these converge to classical bandwidth-optimal algorithms like Ring~\citep{taccl,wang2020blink}, which NCCL implements~\citep{hu2025demystifying}. We provide an in-depth discussion of related work in Appendix~\ref{sec:more-related}.
    \section{\sysname: A Straggler-Aware \allreduce{} Algorithm}
\label{sec:algorithm}
We now present \sysname, a novel algorithm to speed up \allreduce{} in the presence of stragglers. 

\myparab{$\alpha$-$\beta$ cost model.}
\allreduce{} algorithms are typically analyzed using the $\alpha{-}\beta$ model of collective communication~\citep{alphabeta, thakur2005optimization, taccl, won2023astra, wang2025simai}. 
As per this model, sending a message of $s$ bytes takes $\alpha + s \beta$ time, where $\alpha$ is the fixed startup cost per message (independent of message size) and $\beta = \frac{1}{bandwidth}$ is the per-byte transmission cost.

Collective algorithms perform a series of data transfers among GPUs to achieve the desired operation. These algorithms specify 
a \emph{schedule} that dictates which GPUs exchange data in each discrete time step, called a \emph{round}. 
The $\alpha$ cost of an algorithm corresponds to the number of rounds, and the $\beta$ cost corresponds to the total number of 
serialized bytes sent (number of rounds $\times$ bytes sent per round).

The goal of a collective algorithm is to provide a schedule that minimizes the total $\alpha{-}\beta$ cost.  
 CCLs use $\alpha$-optimal algorithms for small
buffers, when latency dominates, and $\beta$-optimal algorithms for larger buffers, when bandwidth becomes the bottleneck (as only $\beta$ cost scales with 
buffer size). Different algorithms affect the \emph{coefficients} on $\alpha$ and $\beta$ while the $\alpha$ and $\beta$ constants come from the hardware. 
Bandwidth-optimal algorithms  (\eg{} Ring) bound $\beta$ cost 
at ${\sim}2s\beta$ regardless of the number of GPUs.
~\sysname{}, like standard \allreduce algorithms used in distributed ML, focuses on optimizing $\beta$ cost because today's large models incur large buffers of many MB to several GB.

\myparab{Key insights.}
In standard \allreduce algorithms, all GPUs must wait for the final GPU, the straggler, before starting the collective (GPU 0 in Fig.~\ref{fig:straggler-diagram}). In contrast, \sysname uses this delay to perform a \reducescatter (Fig.~\ref{fig:collectives}) among non-straggler GPUs. 
By productively using the straggler's delay, \sysname reduces the data transfer cost once the straggler is ready.

Once the \reducescatter is complete, \sysname initiates a schedule of data transfers among all GPUs (straggler and non-stragglers) to complete the \allreduce. 
However, even with the precondition, designing an algorithm faster than $\beta$-optimal baselines is challenging because the precondition introduces inherent asymmetry (straggler vs. other ranks) that is difficult to parallelize. 
Thus, \sysname{} designs communication to maximize parallelism in every round, achieving faster \allreduce{} schedules than baselines.
While computing efficient schedules is known to be combinatorially hard~\citep{taccl}, \sysname{} uses symmetry-breaking to quickly find them in polynomial time.
For the desired multi-GPU setup, \sysname is run once offline to generate \allreduce{} schedules, which are then implemented to run in real-time without modification. 

\sysname is agnostic to the cause of the straggler or detection method (\S\ref{sec:eval}). Instead, it is a \emph{fundamental parallel algorithm that communicates $2\times$ fewer bytes} than the known lower bound for \allreduce{} during exposed communication in settings where overlap is possible, while retaining strong worst-case performance.
We now describe the runtime environment \sysname relies on.

\myparab{GPU cluster topology.} Like classical \allreduce algorithms (\eg Ring, Recursive Halving/Doubling~\citep{thakur2005optimization}), \sysname{} targets the \textbf{scale-up domain}---nowadays extending to 10-100s of GPUs~\citep{nvl72}---due to its homogeneous, any-to-any topology. 

\myparab{GPU-GPU connectivity.} We assume each GPU has a single connection to the inter-GPU network, as is typical in modern switched scale-up domains~\citep{nvlink}. Thus, a GPU can fully utilize bandwidth by sending data to one peer at a time (confirmed empirically in \S\ref{sec:eval}). While multidimensional topologies~\citep{googletpuv4} enable using multiple links in parallel, this proportionally reduces the available per-link bandwidth in typical NVSwitched configurations (as we show in \S\ref{sec:eval}).
\setlength{\textfloatsep}{8pt} 
\begin{algorithm}[!t]
    \caption{\sysname{} Schedule Generator for \allreduce{} (Power-of-2 World Size)}
    \label{alg:stragglar-appendix}
    \KwIn{$n = 2^k$ GPUs, with rank $\sigma = n{-}1$ as the straggler (without loss of generality)}
    \KwOut{\texttt{schedule} of chunk exchanges completing \allreduce{} in $n{-}2 + \log n$ rounds}
    \textbf{Initialization:} rank $g < n{-}1$ holds partially reduced chunk $c_g$; straggler $\sigma$ holds none.\\
    $A$: \texttt{\{\}}, dictionary of active chunks to the set of GPUs that hold it \\
    \texttt{schedule}: \texttt{[]}, list of sets of matchings over rounds to return

    \For{round $r = 0$ \KwTo $n{-}2 + \log n$}{
        \If{$r < n{-}1$}{
            \textbf{Match} rank $r \leftrightarrow \sigma$: exchange chunk $c_r$; mark $r$ and $\sigma$ as unavailable
        }
    
        \uIf{$0 < r < \log n$}{
            \textbf{Match} rank $r{-}1 \to$ rank $r{-}1{+}\log n$: send chunk $c_{r{-}1}$\\
            Each rank with a reduced chunk sends to any rank $g > 2(\log n{-}1)$ without a chunk
        }
        \Else{
            $P_r \gets$ available ranks holding oldest active chunk $\min_{c} A$

            $Q_r \gets$ available ranks holding other active chunks in $A \setminus \{ \min_{c} A \}$\\
            \tcp{Handle ranks in the critical window}
            \For{$g = r{+}1$ to $r{+}\log n$}{
                Select $c = \min\{c' \mid g \text{ lacks } c',\; \forall h \in A[c'],\; h \notin [r{+}1,\, r{+}\log n{-}1]\}$\\
                \textbf{Match} rank $g \leftrightarrow h$; $g$ sends $c$; $h$ sends $c'$; mark $g$ and $h$ as unavailable
            }
        }

       \textbf{Match} $p_i \leftrightarrow q_i$ for $p_i \in P_r, q_i \in Q_r$ to exchange their active chunks

        Append a list of round $r$'s matchings to \texttt{schedule}

        \If{$r \ge \log n$}{
            $A[c_{r-\log n}] \gets \emptyset$  \tcp*{$c_{r - \log n}$ is now fully propagated}
        }

        \If{$r < n{-}1$}{
            $A[c_r] \gets \{r\}$ \tcp*{Add newly active chunk to $A$}
        }
        For other active chunks $c$ exchanged: $A[c] \gets A[c] \cup \{\text{receivers of $c$ in this round}\}$
    }
\end{algorithm}
\setlength{\textfloatsep}{8pt}

\subsection{Algorithm Design}
\vspace{-1em}

\label{sec:alg-design}
We consider a setting with $n$ ranks, $0,{\ldots},n{-}1$. Each rank has a buffer of $s$ bytes that is evenly divided into $n{-}1$ chunks, each of size $\frac{s}{n-1}$ and indexed as $c_{0},{\ldots},c_{n-2}$. Without loss of generality, we assume rank $n{-}1$ is the straggler in this section; however, by symmetry, the algorithm applies regardless of which rank is the straggler (as further shown by our evaluation in \S\ref{sec:eval}).

\myparab{Precondition and postcondition.} \sysname uses a precondition where the $n{-}1$ non-straggler ranks have completed a \reducescatter among themselves, ideally overlapped with the delay of the straggler. This is feasible, since \reducescatter is inherently $2\times$ faster than \allreduce and is efficiently implemented by CCLs~\citep{thakur2005optimization}. We refer to the reduced chunks after this step as \textbf{partially reduced} because they contain all but the straggler's data. \sysname computes an optimized communication schedule to transition from this precondition to the standard \allreduce postcondition, where all ranks, including the straggler, possess a fully reduced buffer. 

\begin{figure}[t]
    \centering
    \begin{subfigure}[t]{0.485\textwidth}
        \centering
        \includegraphics[width=\textwidth]{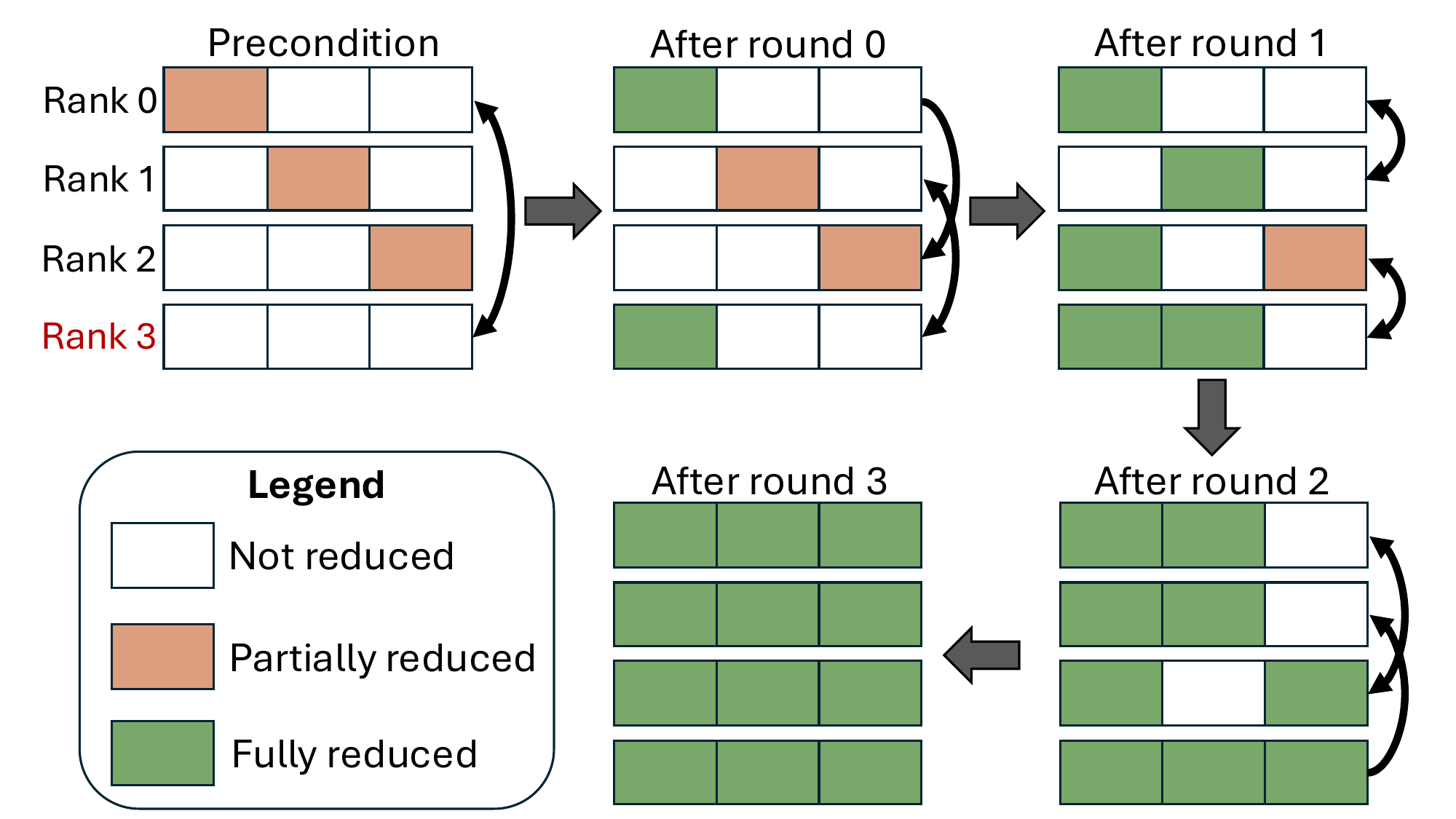}
        \caption{Example \sysname{} schedule for 4 GPUs.}
        \label{fig:schedule}
    \end{subfigure}
    \hfill
    \begin{subfigure}[t]{0.505\textwidth}
        \centering
        \includegraphics[width=\textwidth]{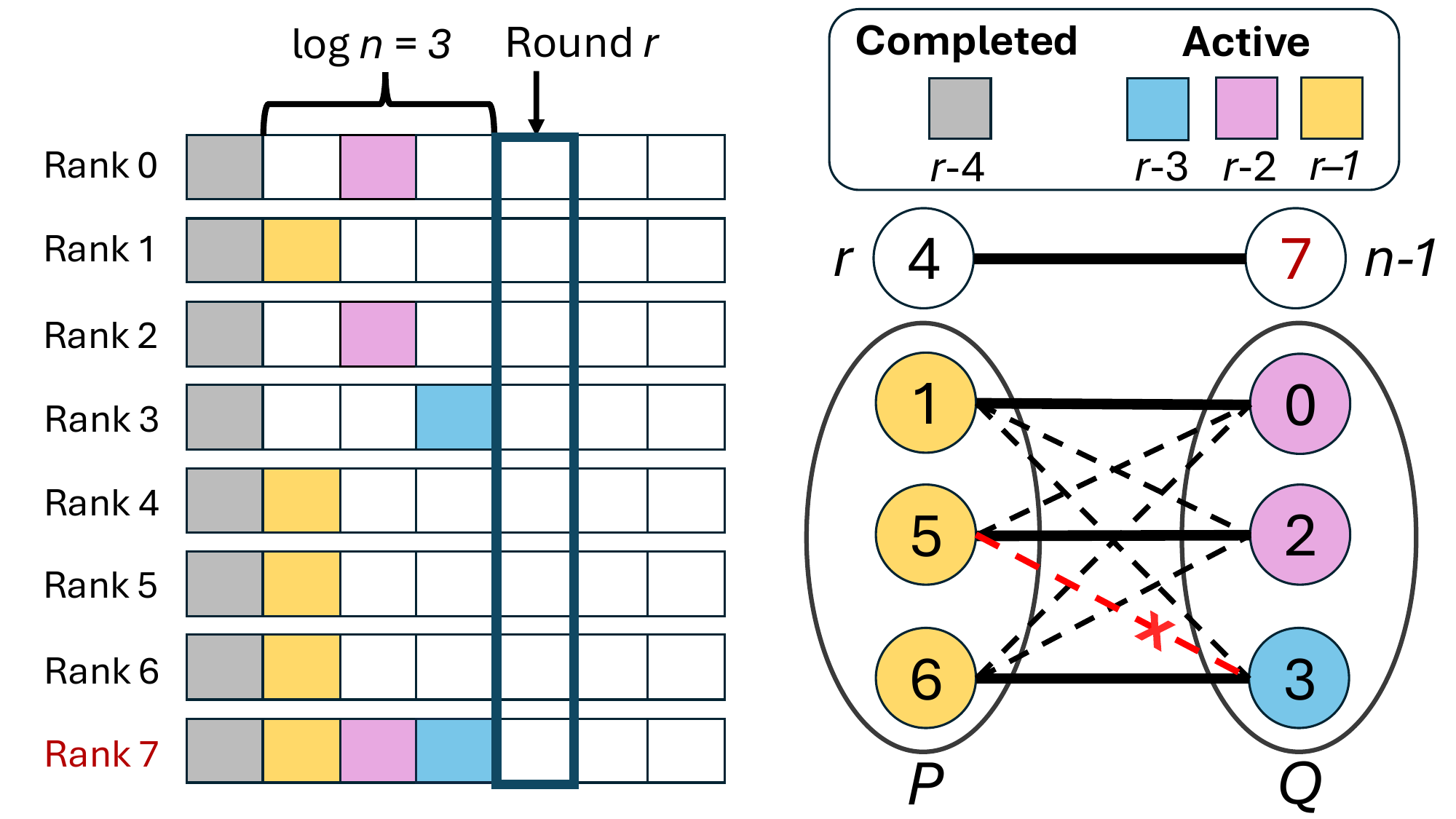}
        \caption{Example matching process, round $r{=}4$ with 8 GPUs.}
        \label{fig:matching}
    \end{subfigure}
    \caption{Algorithm design (straggler in red). In (b), ranks 3 and 5 cannot match: 5 is in the critical window and 3’s chunk was just unlocked. Pairing them would prevent 3’s chunk from doubling.}
    \label{fig:algorithm}
\end{figure}

The communication schedule consists of multiple rounds, each represented by a set of matchings, similar to some broadcast algorithms~\citep{bar2000optimal}. In each round, matchings specify the pairs of ranks that communicate and which data chunks they exchange simultaneously. Each rank participates in exactly one matching per round. Since we fix the chunk size per round, the only lever to minimize communication time is to reduce the number of rounds. At the very least, $n{-}1$ rounds are required to transmit all $n{-}1$ chunks. \sysname{} completes the \allreduce{} in $n{+}\log n{-}2$ rounds, yielding a $\beta$ cost of $\tfrac{n+\log n-2}{n-1} s \beta$, which is less than the known lower bound of $2\tfrac{n-1}{n} s \beta$ (see \S\ref{sec:complexity}). Alg.~\ref{alg:stragglar-appendix} summarizes the schedule computation when $n$ is a power-of-two, with modifications for non-power-of-two cluster sizes in \S\ref{sec:nonpow2}. Fig.~\ref{fig:schedule} visualizes the \sysname{} schedule for $n{=}4$. 

\myparab{Straggler pairings.} In round $r$, rank $r$ pairs with the straggler to fully reduce $c_r$. After $n{-}1$ rounds, the straggler’s buffer is fully reduced and each chunk $c_i$ has been fully reduced on at least one non-straggler rank. Hence, $c_r$ can propagate only from round $r{+}1$ onward, once fully reduced.

\myparab{Phase 1.} In the first $\log n$ rounds, every rank obtains a fully reduced chunk, guaranteed by two rules:
\begin{compactitem}
\item In round $r \in [1, \log n{-}1]$, rank $r{-}1$ sends chunk $c_{r{-}1}$ to rank $r{-}1{+}\log n$.
\item Any other rank with a fully reduced chunk can send it to any other rank $g > 2(\log n{-}1)$.
\end{compactitem}

\myparab{Phase 2.} At the start of round $r = \log n$, every rank has exactly one fully reduced chunk.
Exactly one rank, $r{-}1$, has $c_{r-1}$. Two ranks have $c_{r-2}$ because it was first
fully reduced in round $r{-}2$ and then transmitted to another rank in round $r{-}1$. By the same pattern, four 
ranks have $c_{r-3}$, \etc{}

\myparab{Definition.} An \textbf{active chunk} $c_{j}$ is a chunk that has been fully reduced with the straggler by round $r > j$, 
but has yet to propagate to all $n$ ranks.

As we prove in \S\ref{subsec:proofs}, every non-straggler possesses \emph{exactly one} active chunk at 
any point in time. \sysname{} aims to propagate active chunks to all ranks as quickly as possible. 
Each active chunk requires a minimum of $\log n$ rounds to propagate fully, by doubling in each round, so
an active chunk $c_{j}$ is \textbf{due} for full propagation to all ranks by round $j + \log n$.

Algorithm~\ref{alg:stragglar-appendix} enables every active chunk to double in every round 
by ensuring that every rank transmits its active chunk and receives a different active chunk, with the oldest active chunk expiring in each round.
We model each round as a bipartite matching problem between two disjoint sets of non-straggler ranks: $P_{r}$, ranks whose active chunk 
in round $r$ is $c_{r{-}\log{n}}$ and $Q_{r}$, the others. By the doubling property, it follows that $|P_{r}| = 2^{\log{n}{-}1} = \tfrac{n}{2}$ 
and thus $|Q_r| = n{-}1{-}|P_r|= \tfrac{n}{2}{-}1$.

This allows us to define an invariant: \emph{For every round $r \in [\log n, n{-}2]$, rank $r \in P_{r}$.}
This guarantees the $\log n$ propagation deadline, by ensuring that ranks receiving a new active chunk (by pairing with the straggler) have already received the chunk due in that round ($c_{r{-}\log n}$). 
Thus, we exclude rank $r$ from the matching process since it pairs with the straggler, enabling $|P_r| = |Q_r| = \frac{n}{2}{-}1$.

Now, we have two equal-sized, disjoint sets of ranks, $P_r\setminus\{r\}$ and $Q_r$. However, arbitrary matchings between the two sets
risk violating the invariant \emph{in the future}. In particular, ranks in the next $\log n$ rounds (the \textbf{critical window}) 
cannot receive chunks that would remain active when they must pair with the straggler (Fig.~\ref{fig:matching}). For example, rank $r{+}1$ may only receive $c_j$ with $j \le r{+}1{-}\log n$.
To enforce this, we first match ranks $j \in  [r{+}1, r{+}\log{n}]$ in the critical window with partners outside 
this window that have the oldest available active chunk. After 
finalizing these matchings, the remaining ranks can be paired arbitrarily since every $u \in P_r$ and $v \in Q_r$ 
lack each other's active chunks.

After $n{-}1$ rounds, every chunk has been fully reduced, and the straggler has fully reduced its buffer.
To fully propagate remaining chunks quickly, the straggler can be paired with arbitrary ranks, but always sends the final chunk. (This enables the straggler 
to be added to $Q_r$ so that $|P_{r}|=|Q_{r}|=\tfrac{n}{2}$.)

\subsection{Communication Complexity}
\label{sec:complexity}
\myparab{Theorem 1.} \emph{\allreduce{} schedules generated by \sysname{} complete in $n{+}\log n{-}2$ rounds.}

We provide the proof in \S\ref{subsec:proofs}. Intuitively, 
each chunk $c_r$ propagates within $\log n$ rounds of being fully reduced.  
The final chunk $c_{n-1}$ only requires $\log n{-}1$ rounds to propagate, as the straggler can also help transmit it, thereby 
achieving $n{+}\log n{-}2$ total rounds. Achieving this bound 
requires matching ranks optimally for the current round without 
compromising on future propagation, where special care must be taken due to the pre-determined straggler pairings.

\begin{wraptable}{r}{0.55\linewidth}
\vspace{-1em} 
\centering
\scriptsize
\renewcommand{\arraystretch}{0.9}
\begin{tabular}{@{}lcc@{}}
\toprule
Algorithm & Latency & Bandwidth \\
\midrule
Ring       & $2(n{-}1)\alpha$ & $\tfrac{2(n-1)}{n}s\beta$ \\
RHD       & $\pmb{2(\log n)\alpha}$ & $\tfrac{2(n-1)}{n}s\beta$ \\
\sysname{} & $(n{+}\log n{-}2)\alpha$ & $\pmb{\tfrac{n{+}\log n{-}2}{n{-}1}s\beta}$ \\
\bottomrule
\end{tabular}
\caption{\allreduce{} complexity with straggler delays.}
\label{tab:allreduce-cost}
\end{wraptable}

When the straggler is delayed long enough to mask the \reducescatter{} 
precondition (supported by Fig.~\ref{fig:intro-motivation}), \sysname{'s} exposed communication time consists of $n{+}\log n{-}2$ rounds with 
$\tfrac{s}{n-1}$ bytes sent per round. Thus, the total time taken is 
\begingroup
\setlength{\abovedisplayskip}{5pt}
\setlength{\belowdisplayskip}{5pt}
\[
T_{\text{SAR}} = (n + \log n - 2)\,\alpha
  + \frac{n + \log n - 2}{n - 1}\, s \beta.
\]
\endgroup
Under these conditions, \sysname{} achieves much lower $\beta$ cost than today's known bandwidth-optimal lower bound (see Table~\ref{tab:allreduce-cost}), and this 
advantage only grows as $n$ (number of GPUs) scales. The scaling behavior of known $\beta$-optimal algorithms is 
$\lim_{n \to \infty} \tfrac{2(n{-}1)}{n}s\beta = 2s\beta$, whereas \sysname{} achieves $\lim_{n \to \infty} \tfrac{n{+}\log n{-}2}{n{-}1}s\beta = s\beta$, 
implying $2\times$ speedups in large-scale settings with a straggler. Even in the worst case, where none of the \reducescatter{} 
precondition can be overlapped (\eg no stragglers), \sysname{} remains competitive: 
since a \reducescatter{} with $n{-}1$ ranks only incurs $\tfrac{n{-}2}{n{-}1}s\beta$ cost, \sysname{} achieves $\lim_{n \to \infty} \tfrac{n{-}2}{n{-}1}s\beta + \tfrac{n{+}\log n{-}2}{n{-}1}s\beta = 2s\beta$. While $\alpha$ cost is typically negligible 
for large buffers, \sysname{} scales better in $\alpha$ cost than Ring but more poorly than RHD.
     \section{Experiments}
 \label{sec:eval}
We implement the \sysname algorithm to compute schedules, as well as a CUDA runtime that executes these schedules in multi-GPU setups. 

\myparab{Detecting stragglers.} One advantage of \sysname{} is that its initial \reducescatter can be eagerly executed as soon as the first $n{-}1$ ranks are ready. If there is sufficient delay between the slowest rank and these $n{-}1$ ranks, \sysname{} achieves its ideal performance; otherwise, performance falls in the range between the ideal and worst-case bounds discussed in \S\ref{sec:algorithm} and \S\ref{subsec:scaling}. To consistently support ideal-case performance, the runtime can leverage online straggler detection tools~\citep{zhao2024adapcc} and pass the straggler's rank to the backend for conditional execution of the appropriate schedule. In our end-to-end experiments (\S\ref{subsec:e2e}), we fix the rank that \sysname assumes to be the straggler (by profiling the workload ahead of time and picking a likely straggler rank, see \S\ref{subsec:e2e}). This stress-tests \sysname, as there are many iterations in which the algorithm encounters its worst-case performance (\ie no precondition overlap), when a different rank is the straggler or there is no straggler at all.

\myparab{Schedule generator.} We implement \sysname in Python to generate \allreduce schedules. The algorithm takes the number of GPUs as input and outputs a schedule with matchings, specifying for each round which GPUs communicate and which chunks to send/receive. Schedule generation is run once offline and is fast---\sysname computes the schedule for a 256-GPU cluster in \textless 1.04 seconds.

\myparab{Runtime.} We implement the \sysname{} schedules using the NCCL Point-to-Point (P2P) API~\citep{nccl} (also used in prior work~\citep{hwang2023tutel}), for GPU communication, combined with custom CUDA kernels for reduction. After completing \texttt{ncclReduceScatter()} among non-straggler GPUs, we invoke the runtime to execute the \sysname{} schedule. We package both \reducescatter{} and \sysname{} into an API that has the same functionality as \texttt{ncclAllReduce()}.

\myparab{Baselines.} 
Given the homogeneous, switched connectivity of modern scale-up networks, we compare \sysname{} to the strongest bandwidth-optimal \allreduce{} algorithms for this setting.
Our baselines are \textbf{(1) Ring}~\citep{patarasuk2009bandwidth, nccl-ring} and \textbf{(2) Recursive Halving/Doubling (RHD)}~\citep{bruck1994efficient}, which are the 
gold-standard bandwidth-optimal algorithms in homogeneous networks~\citep{hu2025demystifying}; \textbf{(3) MSCCL}~\citep{cowan2023mscclang}, the only recent work 
we are aware of to synthesize a new algorithm for scale-up domains; and \textbf{(4) Broadcast}, a naive straggler-aware baseline. 
The MSCCL baseline uses their \emph{allpairs} algorithm: a one-round \reducescatter{} where each GPU splits its bandwidth across all peers, followed by a one-round mirror \allgather{}. This achieves the same $2s\beta$ bandwidth cost as classical algorithms but reduces latency to $2\alpha$ on switched networks. 
The Broadcast baseline is straggler-aware, as non-stragglers complete an \allreduce{} during the straggler delay and then finish with a pairwise-exchange broadcast once the straggler arrives. More details on baselines can be found in \S\ref{sec:baselines}. For fair comparison of the algorithmic contribution, we implement baselines using the NCCL P2P API and the same CUDA compute kernels as \sysname.

 \subsection{Benchmarking \sysname on Multi-GPU Scale-Up Domains}
 \begin{figure}[t]
    \centering
    \includegraphics[width=\linewidth]{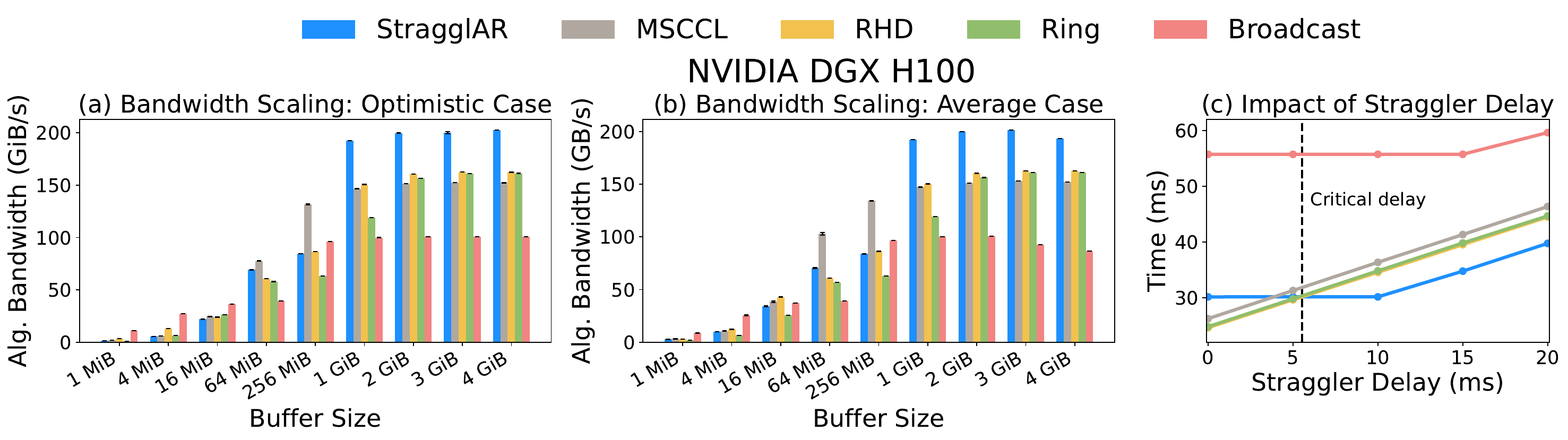}
    \includegraphics[width=\linewidth]{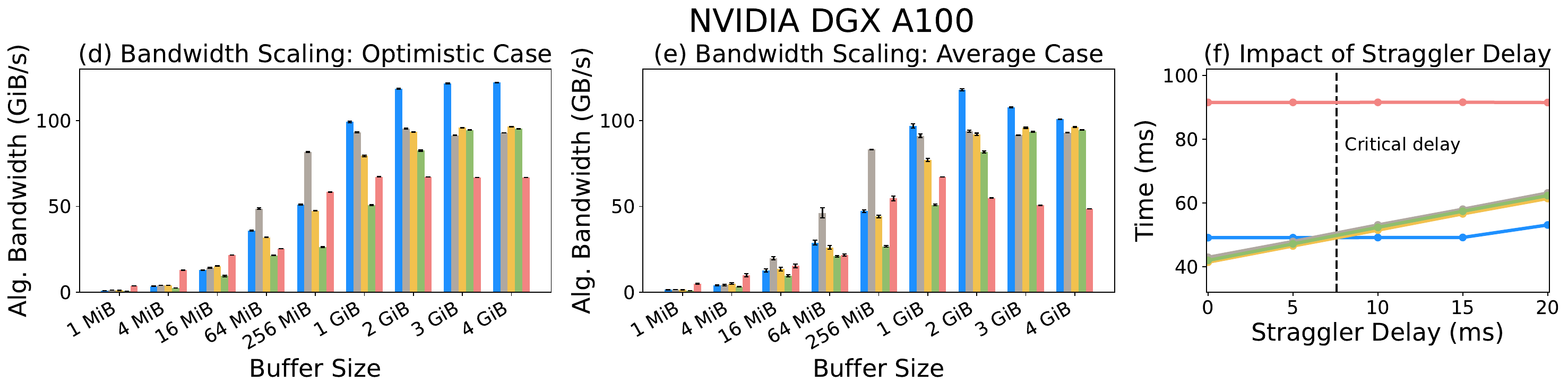}
    \caption{Benchmarking \allreduce{} performance for different algorithms on NVIDIA DGX H100 (a-c) and A100 (d-f) servers. (a), (d) show the optimistic use case, by assuming 
    \reducescatter{} can complete within the straggler delay. (b), (e) use a straggler delay of 9.45 ms, the average from the ML workloads in Fig.~\ref{fig:intro-motivation}. 
    (c), (f) fix the buffer size at 4 GiB and vary the straggler delay.}
    \label{fig:8gpu}
 \end{figure}

 \myparab{Setup.} We run experiments on several multi-GPU testbeds: (1) NVIDIA DGX H100, 8$\times$80GB H100 GPUs connected by NVLink 4.0 (450 GB/s P2P) and NVSwitch 3.0; (2) 
 NVIDIA DGX A100, 8$\times$80GB A100 GPUs connected by NVLink 3.0 (300 GB/s P2P) and NVSwitch 2.0; 
 (3) a node of the Perlmutter supercomputer~\citep{nersc_perlmutter_architecture},
 with 4$\times$40GB A100 GPUs fully connected (mesh) by NVLink 3.0 for 100 GB/s P2P (results in \S\ref{sec:perlmutter}). 
 (1) and (2) are VMs obtained through RunPod~\citep{runpod2025}. We use CUDA events on GPU rank 0 to measure the runtime.

 \myparab{Varying buffer size.} We first assume the straggler delay can mask the \reducescatter{} precondition and measure runtime from this point to capture performance for ideal use cases of our algorithm.
 (For Broadcast, we assume the entire \allreduce{} precondition has completed.) Following the standard for benchmarking \allreduce{} performance, we evaluate the communication time as we scale the buffer size
 in powers of 2 (in bytes). NCCL behaves unpredictably when the P2P data transfer size (\ie chunk size)
 is not a multiple of 4 KiB (page size), so we pad buffers for \sysname{} to ensure the chunk size is the lowest multiple of 
 4 KiB \emph{greater} than $\tfrac{s}{n-1}$ (for baselines, the chunking scheme inherently ensures this when $s$ is a power of 2). For each buffer size, we run 50 iterations per algorithm and report the mean, with error bars for the standard error of the mean.

 Similar to how NCCL reports performance~\citep{nccl_tests}, we show the \emph{algorithmic bandwidth}---buffer size divided by  
 collective communication time (\ie{} runtime normalized by input size)---in Fig.~\ref{fig:8gpu} (a,d). For some smaller buffer sizes, RHD, Broadcast, and MSCCL
 outperform \sysname{}, which still outperforms Ring; this matches expected results from $\alpha{-}\beta$ costs since $\alpha$ costs dominate
 for small buffers. However, \sysname{} is consistently the fastest for the larger 
 buffer sizes (which scale with model size), specifically 8.3\% faster on the 4-A100 Perlmutter node and >25\% faster on the 8-H100 and 8-A100 DGX servers. The outlier at 256 MiB likely stems from NCCL’s internal tuning in the 64-512 MiB range, where internal protocol changes cause 
 unexpected performance, as confirmed by our own \texttt{nccl-tests} profiling (\S\ref{subsec:nccl-tests}) and prior work~\citep{xu2025autoccl,hu2025demystifying}.

 To simulate stragglers, we idle GPU rank $n{-}1$ 
 for the specified number of clock cycles. Simultaneously, we begin the \reducescatter{} for the other $n{-}1$ 
 GPUs for \sysname{} (similarly, \allreduce{} for Broadcast). We conduct similar experiments as above, but instead impose the 
 average delay of 9.45 ms observed in our Llama-3.2 fine-tuning experiments in Fig.~\ref{fig:motivation}. 
 Fig.~\ref{fig:8gpu} (b,e) shows that \sysname{} improves the runtime under typical straggler delays in distributed ML with larger buffer sizes, with 
 average-case performance closely matching ideal performance. 
 For the 4 GiB buffer, \sysname{'s} performance declines slightly from the ideal case because the \reducescatter{} for this 
 buffer size cannot be fully overlapped with the 9.45 ms straggler delay.

 \myparab{Varying straggler delay.} We vary the straggler delay by adjusting the number of clock cycles for the sleep kernel, and run all 
 algorithms on the largest buffer size (4 GiB) as a stress-test of the precondition overlap. 
 Fig.~\ref{fig:8gpu} (c,f) captures the total time from when the \allreduce{} is called on non-straggler GPUs to when it completes.
 It shows the critical delays of 5.53 ms and 7.57 ms for the DGX H100 and A100, respectively, after which \sysname outperforms baselines. The critical delay is higher for the DGX A100 since a \reducescatter{} takes longer due to its lower P2P bandwidth.
 The critical delay also depends on buffer size, as shown in Fig.~\ref{fig:reducescatter}, which plots the runtime of the \reducescatter{} precondition on the DGX H100. 
 While the shaded region in Fig.~\ref{fig:reducescatter} captures the straggler delays at which \sysname{} achieves its full theoretical gains, \sysname{} still outperforms baselines when some, 
 but not all, of the \reducescatter{} is overlapped with the straggler delay. For example, the critical delay for a 4 GiB buffer in Fig.~\ref{fig:8gpu}c is less than 
 the \reducescatter{} time for 4 GiB. Thus, \sysname{} enables speedups even if the straggler delay is less than the \reducescatter{} precondition, but longer than the critical delay. 
 As we show in \S\ref{subsec:scaling}, \textbf{the critical delay approaches zero as the size of the GPU cluster increases} due to \sysname{'s} efficient scaling.
\vspace{-0.5em}
 \subsection{End-to-end Evaluation on ML Workloads}
\label{subsec:e2e}

We run experiments on DGX A100 VMs to assess \sysname{'s} impact on end-to-end training time with data-parallel fine-tuning of three  
popular LLMs: Llama-3.2-3B~\citep{grattafiori2024llama}, Phi-3-mini-3.8B~\citep{abdin2024phi}, and 
Qwen-2.5-3B~\citep{team2024qwen2}. We compile our \allreduce{} implementations
into a custom package and configure PyTorch's \allreduce{} to call this backend.
In each VM, we first profile the workload with standard PyTorch tools and identify persistent stragglers, \ie ranks that are most likely to be delayed based on data from prior runs.
We use this approach both because persistent stragglers are common according to prior work~\citep{lin2025understanding,jiang2024megascale} and our own experiments (\S\ref{subsec:persistence}), and because it stress-tests 
\sysname to encounter both ideal and worst-case conditions (when a different/no rank is the straggler).
Then, we pass the selected rank to the backend and train the model for 100 iterations (batch size of 32).
Because the buffer sizes of these models are $\sim$4 GiB, we compare \sysname{} to the Ring algorithm, which is optimal at large buffer sizes (Fig.~\ref{fig:8gpu}a).
We show the end-to-end speedups in Table~\ref{tab:e2e}.

\begin{wraptable}{r}{0.58\textwidth}
\vspace{-1em}
\centering
\small
\renewcommand{\arraystretch}{0.95}
\begin{tabular}{lccc}
\toprule
Model & Speedup & Straggler & GPU hrs. \\
      & (\%)   & persistence (\%) & saved/day \\
\midrule
Llama-3.2-3B    & 4.75 & 90 & 9.12\\
Phi-3-mini-3.8B & 4.43 & 95 & 8.51\\
Qwen-2.5-3B     & 2.39 & 77 & 4.59 \\
\bottomrule
\end{tabular}
\caption{End-to-end training speedups over Ring, and GPU-hours saved per day on an 8-GPU server.}
\label{tab:e2e}
\end{wraptable}
The end-to-end speedup depends on several factors: how often the actual straggler was 
the rank we passed to the backend, the straggler's delay, and the fraction of overall time spent on \allreduce{}. 
For Qwen-2.5-3B, gains are smaller because the straggler on that VM was less persistent, leading our algorithm 
to encounter its worst-case scenario (\ie no \reducescatter overlap) more often. Even then, \sysname{} consistently 
shows end-to-end gains, as its worst-case performance nearly matches baselines while its upside is much higher (Fig.~\ref{fig:e2e-projections}). 
\sysname does not require online straggler detection with dynamic stragglers, as eager conditional execution of schedules based on the first $n{-}1$ ready ranks 
means at worst (\ie no straggler delay), \sysname's performance closely matches baselines. These speedups 
translate to 9.12 GPU-hours saved per day on an 8-GPU server (Table~\ref{tab:e2e}).

\subsection{Scaling Characteristics of \sysname}
\label{subsec:scaling}
\begin{figure}[t]
    \centering
    \begin{subfigure}[t]{0.4\linewidth}
        \centering
        \includegraphics[width=\textwidth]{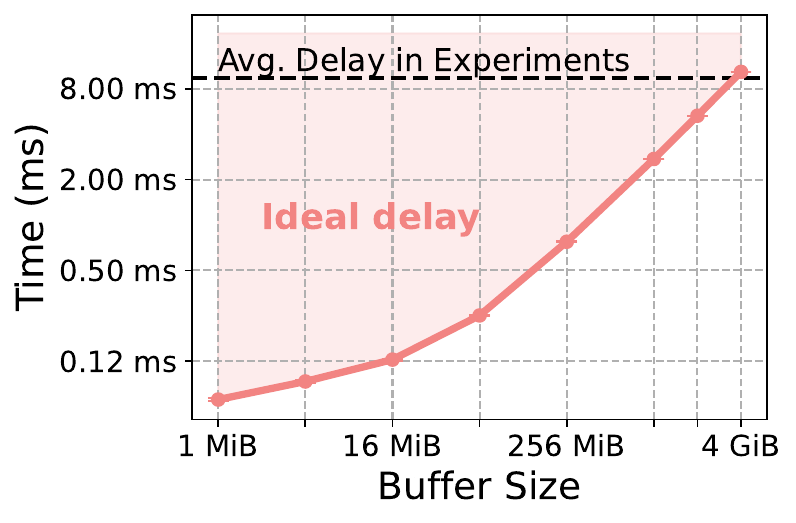}
        \caption{Ideal straggler delay (DGX H100).}
        \label{fig:reducescatter}
    \end{subfigure}
    \begin{subfigure}[t]{0.45\linewidth}
        \centering
        \includegraphics[width=\textwidth]{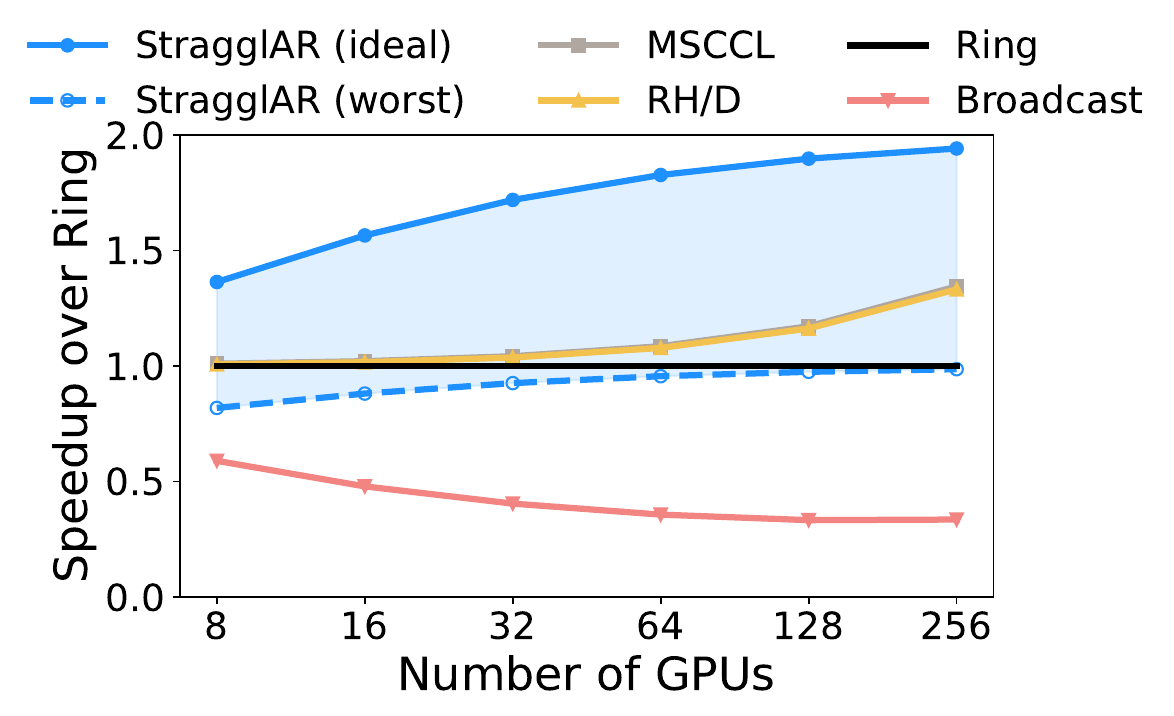}
        \caption{Speedups over Ring as cluster size scales.}
        \label{fig:simulation}
    \end{subfigure}
    \caption{(a) \reducescatter{} time over buffer sizes, capturing 
    the range of straggler delays where \sysname achieves full theoretical guarantees. (b) \allreduce{} performance scaling with the analytical model for a 1 GiB buffer. The shaded region is the range of \sysname{'s} possible speedups.}
    \label{fig:scaling}
\end{figure}

To assess scaling to the largest practical scale-up domains (256 GPUs), 
we employ the same approach as prior work: using the popular and empirically validated analytical network model to simulate performance 
on clusters larger than 8 GPUs~\citep{won2023astra, wang2025simai, gui2025accelerating,won2024tacos}, as we lack access to 
hardware like NVIDIA's GB200~\citep{nvl72}.
The analytical simulator uses the $\alpha{-}\beta$ model to predict performance. 
We use $\alpha$ = $3\mu$s based on latency profiling studies of recent NVLinks~\citep{nvlink_latency} and $\beta$ as the inverse of 450 GB/s, the P2P bandwidth on DGX H100. 
In these simulations, we capture the \emph{entire range} of performance supported by \sysname{}, from the worst case where none of the initial \reducescatter{} 
can be overlapped to the ideal case where the straggler delay enables full overlap (see Fig.~\ref{fig:reducescatter} for ideal range).
In Fig.~\ref{fig:simulation}, both \sysname{'s} ideal and worst-case performance improve with cluster size for realistic, large buffer sizes (1 GiB) in distributed ML. 
By $n=256$,~\sysname{} provides a nearly $2\times$ speedup over the Ring algorithm in straggler settings while being \emph{no worse} even without stragglers.
Thus, incorrect or infeasible straggler detection has minimal impact, as \sysname exhibits competitive bandwidth efficiency even at its worst.
We report end-to-end scaling results in simulation for ML training in \S\ref{sec:e2e-simulation}.

\myparab{Limitations.} To consistently achieve its best-case performance with dynamic stragglers, \sysname{} 
requires conditional execution of schedules based on the $n{-}1$ ranks that are first ready, which can be complex.
We note that libraries like NCCL 
already support conditional execution of different algorithms and protocols based on runtime conditions.
While \sysname{} matches baselines at scale even without a straggler, its performance on smaller clusters 
depends on the critical delay, which varies with the exact GPU P2P bandwidth. Our algorithm also does not 
support odd values of $n$, though such setups are atypical in large-scale ML. Finally, while \sysname can tolerate multiple stragglers, since a straggler is by definition relative, 
it is less effective when many GPUs straggle \emph{simultaneously}; however, this scenario is highly improbable since GPU execution times are continuous variables~\citep{dean2013tail, warraich2025optireduce}.
    \section{Conclusion}
\label{sec:conclusion}
We design \sysname, a parallel algorithm that exploits natural variation in GPU execution times to speed up \allreduce. \sysname{} achieves a $2\times$ speedup over the known lower bound for bandwidth-optimal \allreduce{}, while performing as well in the worst case. By introducing the dimension of \emph{temporal asymmetry}, \sysname{} offers a new paradigm for collective algorithm design. 
    \section*{Acknowledgements}
This research is supported by NSF
Award \#2444537 and ACE, one of the seven
centers in JUMP 2.0, a Semiconductor Research Corporation
(SRC) program sponsored by DARPA. AD is supported by the 
NSF Graduate Research Fellowship, and AVK is supported by 
the Cornell Bowers CIS-LinkedIn Grant.
This research used resources of 
the National Energy Research Scientific Computing Center, a DOE 
Office of Science User Facility supported by the Office of Science of the 
U.S. Department of Energy under Contract No. DE-AC02-05CH11231 
using NERSC award NERSC DDR-ERCAP0032726. We thank Giulia Guidi for 
making supercomputing resources available via the CS 6230 course at Cornell.
    \newpage
    \bibliography{reference}
    
    \appendix
    \newpage
\section*{Appendix}
\section{Extended Related Work}
\label{sec:more-related}
We provide more discussion of related work in this section.

In distributed training and inference, each GPU produces local gradients and activations that must be aggregated across devices. This multi-device aggregation is performed using the \allreduce collective communication primitive, which transmits and reduces data across the inter-GPU network using optimized parallel communication algorithms. The parallelization strategy of an ML job determines which GPUs communicate and how often: data parallelism uses \allreduce regularly to average gradients, while tensor model parallelism~\citep{megatron} invokes \allreduce many times within each model pass to exchange activations. In ML software stacks, collective communication libraries (CCLs) like NCCL implement algorithms for collective primitives, interfacing with ML application libraries (\eg PyTorch~\citep{pytorch}, TensorFlow~\citep{tensorflow}) at one end and network transport protocols at the other. For \allreduce, NCCL chooses between multiple algorithms based on the number of GPUs and the amount of data to be reduced. Since buffer sizes in modern ML workloads tend to be large~\citep{taccl}, CCLs leverage bandwidth-optimal algorithms (\eg Ring) to minimize the \allreduce time.

Most CCLs, including NCCL, adopt a bulk-synchronous model where all GPUs synchronize before the collective, guaranteeing data correctness and enabling efficient parallel communication algorithms that rely on all ranks to implement the collective concurrently. However, performance degrades significantly when one or more GPUs are slow to reach the synchronization point. The slowest GPU (the straggler) dictates overall performance due to tail effects~\citep{warraich2025optireduce, dean2013tail, wang2024towards, gangidi2024rdma, li2014tales}. Even with multiple slower GPUs, there is typically only a single straggler since the probability of multiple GPUs completing exactly simultaneously is extremely small. Straggler delays may stem from hardware issues like thermal throttling and power supply, or runtime factors like network congestion, background noise, and unavoidable skewed compute allocations across GPUs~\citep{wu2024falcon, jiang2024megascale, grattafiori2024llama, xiong2024superbench}. Recent work has highlighted the significant impact of stragglers on datacenter-scale ML jobs~\citep{jiang2024megascale, lin2025understanding}, and our experiments fine-tuning Llama-3.2 (1B/3B) across different hardware setups (Fig.~\ref{fig:intro-motivation}) reveal straggler effects of up to 30 ms even within individual multi-GPU servers.

\begin{table}[h]
    \centering
    \scriptsize
    \label{tab:comparison}
    \begin{tabular}{lccccc}
        \toprule
        \textbf{Method} & \textbf{Domain} & \textbf{Backend} & \textbf{Application} & \textbf{Effect on Loss/Convergence} \\
        \midrule
        \citep{harlap2016addressing} & Scale-out & Custom C++ runtime & DP & No \\
        \citep{karakus2017straggler} & Scale-out & MPI & DP & Yes \\
        \citep{wang2020blink} & Scale-out & NCCL & DP, TP & No \\
        \citep{won2024tacos} & Irregular scale-up & Simulation & DP, TP & No \\
        \citep{zhao2024adapcc} & Scale-out & Custom CUDA runtime & DP, TP, EP & No \\
        \citep{warraich2025optireduce} & Scale-out & Gloo \& custom transport & DP & Yes \\
        \textbf{\sysname (Ours)} & \textbf{Scale-up} & \textbf{NCCL} & \textbf{DP, TP} & \textbf{No} \\
        \bottomrule
    \end{tabular}
    \caption{\sysname \vs prior work (DP=data parallel, TP=tensor parallel, EP=expert parallel). While \sysname targets the scale-up domain, it also applies to switched homogeneous scale-out domains, like in today's rail-optimized topologies for ML.}
    \vspace{-1em}
    \label{tab:related}
\end{table}

\myparab{Straggler mitigation.} 
Some prior works attempt to identify and remove straggling devices or thoroughly investigate root causes~\citep{jiang2024megascale,lin2025understanding}. However, removing stragglers wastes compute capacity, and only applies to severe stragglers rather than the routine compute heterogeneity we observe regularly, even at small scales. Moreover, stragglers arise from many sources, making diagnosis difficult and incomplete~\citep{lin2025understanding}. Recent works obviate straggler delays by approximating or dropping the straggler's data during training~\citep{warraich2025optireduce,harlap2016addressing,karakus2017straggler,recht2011hogwild}. While these approximations and asynchronous techniques can progress without waiting for stragglers, they are limited to data-parallel training and can impact model convergence. In contrast, \sysname is a fundamentally new \allreduce algorithm that ensures accurate, unmodified reductions and is agnostic to the \allreduce use case, making it applicable to both data-parallel training and tensor-parallel training/inference. Like all collective algorithms, \sysname requires participation from all GPUs (including the straggler), but accelerates the collective itself. Other works mitigate stragglers in datacenter-scale ML jobs through systems-level workload rebalancing and runtime strategies~\citep{wu2024falcon,zhao2024adapcc}, improving efficiency by adapting task placement or dynamically selecting among different known collective algorithms based on profiling. For example, our investigation of AdapCC~\citep{zhao2024adapcc} finds that it relies classical Tree algorithms, which sacrifice bandwidth to improve latency. In contrast, \sysname{} is not a runtime controller, but a new bandwidth-efficient parallel algorithm for \allreduce that changes the communication schedule itself to provably reduce communication complexity. It can be integrated with different runtime systems that implement these algorithms.

\myparab{\allreduce{} algorithms.} Collective algorithms are designed based on knowledge of the underlying network topology. 
Inter-GPU networks typically fall into two categories: scale-up (homogeneous high-bandwidth links within a node or rack) and scale-out (heterogeneous or unpredictable bandwidth across multiple nodes or racks). 
Recent works have synthesized new \allreduce{} algorithms for heterogeneous networks with asymmetric bandwidth~\citep{taccl,zhao2024adapcc,wang2020blink,won2024tacos,zhang2024rethinking}. 
However, for the homogeneous networks that define modern scale-up domains, these synthesizers invariably converge to bandwidth-optimal classical algorithms, such as Ring and Recursive Halving/Doubling~\citep{taccl,wang2020blink}. 
Consequently, state-of-the-art CCLs like NCCL continue to implement classical algorithms for collective communication in scale-up domains~\citep{hu2025demystifying}.
We provide Table~\ref{tab:related} to concretely situate our contribution in the space of collective algorithms and straggler mitigation strategies. We note that while 
we focus on the scale-up domain for \sysname, it applies to any homogeneous switched network, including rail-optimized scale-out domains~\citep{nvidia_rail_optimized}.

\myparab{Optimal collective communication.}
Over decades, researchers have implemented collective communication primitives with fast algorithms that are carefully attuned to latency and bandwidth trade-offs. With today's large ML models that require communicating large amounts of data, 
bandwidth-efficient algorithms are of primary focus~\citep{narayanan2021efficient, fei2021efficient}. In homogeneous networks, well-known parallel algorithms, such as Ring, recursive halving/doubling, \etc, enable fast, bandwidth-efficient \allreduce{}
by heavily parallelizing communication in every round of the algorithm~\citep{patarasuk2009bandwidth, rabenseifner2004optimization, rabenseifner2004more, thakur2005optimization, bruck1994efficient, sanders2009two}. 
For heterogeneous networks, recent works synthesize custom collective algorithms by using the hardware topology or parallelization strategy as an input~\citep{taccl, kim2024tccl, cai2021synthesizing, zhao2024forestcoll, laskar2024enhancing, won2024tacos, mahajan2023better}; 
optimizing a collective algorithm to minimize the latency and bandwidth cost, however, is a challenging combinatorial optimization problem that often requires hand-tuned heuristics to constrain the search space~\citep{taccl}.
Further, some approaches optimize \allreduce{} for a specific topology~\citep{jain2010optimal, de2024swing, wang2023topoopt}. Our work extends this line of research by optimizing \allreduce{} 
in settings with a straggler for distributed ML, where bandwidth-efficient algorithms are required due to the large buffer sizes~\citep{narayanan2021efficient, taccl,nccl-ring}.

\myparab{Rebalancing workloads.} Recent works improve the efficiency of distributed ML by optimizing the model parallelization strategy, which can affect data/work sharding across GPUs and the collective communication incurred~\citep{jia2019beyond, zhang2024rethinking, zheng2022alpa, shoeybi2019megatron, rajbhandari2020zero, grattafiori2024llama, narayanan2021efficient, unger2022unity,zhao2024adapcc}. Vanilla parallelization strategies evenly shard or replicate the compute and memory load (\ie model weights, input batches, activations, optimizer states, \etc) across GPUs within a group to achieve speedups from parallelization. This balances the computation and communication load experienced by each worker, but could introduce resource idling when there are slow GPUs in the group. 
Our experiments in Fig.~\ref{fig:motivation} and \S\ref{subsec:persistence} highlight significant straggler effects even with uniform workload placement.
Other works identify severe stragglers and balance the workload dynamically to limit their impact~\citep{harlap2016addressing, wu2024falcon, wang2023fluid, lin2025understanding}.  
\sysname{} directly addresses stragglers at a lower layer of abstraction, within the \allreduce{} kernel, and thus expands on the problems tackled by these works.

\myparab{Sending less data.} Various methods leverage the resilience of stochastic optimization algorithms used in training to reduce the communication volume by selectively communicating data with high statistical value \citep{warraich2025optireduce, li2022near, wang2023cocktailsgd}, synchronizing gradients with varying frequencies~\citep{giladi2023dropcompute, hierarchical_sgd, liu2022communication,karakus2017straggler, yang2020mitigating, yang2020mitigating}, or sending lower-precision or compressed data~\citep{alistarh2017qsgd, bernstein2018signsgd, wang2022fine, wang2024zero++, wang2023cocktailsgd, lu2022maximizing, tang20211, renggli2019sparcml, m2021efficient, wang2023hi}.
~\sysname{} is orthogonal to these works because it optimizes the collective algorithm itself (\ie{} \emph{exactly how} the data is transmitted as opposed to \emph{how much}) and can operate given any buffer size, including after compression. 
Further, because \sysname{} directly modifies the collective algorithm, as opposed to the data that is sent or the frequency of synchronization, it also supports \allreduce{} for tensor parallelism, which is used in both training and inference~\citep{megatron}.

\myparab{Compute-communication overlap.} Recent works propose fine-grained overlap of compute and communication to mitigate synchronization delays and reduce communication overheads in distributed ML~\citep{ismayilov2023multi, jangda2022breaking, wang2023mgg, nvshmem, punniyamurthy2024optimizing, torchtitan, pati2024t3, chen2024centauri, wang2022overlap}.
Although this approach may be useful for small, irregular communication patterns, such as expert-parallel \alltoall~\citep{liu2024deepseek}, 
it is less useful with large data buffers that require aggregation and are bottlenecked by network bandwidth~\citep{dryden2018aluminum,patarasuk2009bandwidth}. Thus, bulk data transfers, like gradient synchronization in data-parallel training or activation aggregation in tensor-parallel training with large batch sizes, rely on bandwidth-optimal collective communication.
Further, these works consider overlapping compute and communication at higher layers of abstraction, but rely on the same algorithms to implement the collective communication. \sysname instead brings the idea of overlap to the communication algorithm itself, and uses this to directly address the straggler problem. 
Other works propose entirely abandoning synchronization barriers and relying on resilience of stochastic gradient descent to enable convergence during training~\citep{nabli2023textbf, tyurin2024shadowheart, su2022gba, tyurin2024optimal, de2023epidemic,recht2011hogwild}. 
However, these approaches impact model convergence, limiting applicability, and do not generalize to tensor-parallel training and inference, which require accurate and complete reductions.

\section{Persistent Stragglers}
\label{subsec:persistence}
While recent works~\citep{jiang2024megascale,lin2025understanding} have shown persistent stragglers in ML jobs with thousands of GPUs, we show that even smaller multi-GPU jobs encounter such stragglers. We run Llama-3.2 fine-tuning jobs in three hardware environments: (1) 4 A100s with 40GB memory and NVLink 3.0 fully-connected mesh network in the Perlmutter supercomputer~\citep{nersc_perlmutter_architecture}, (2) 4 A100s with 80GB memory in RunPod~\citep{runpod2025}, and (3) 8 A100s with 80GB memory in RunPod. Both RunPod servers use NVLink 3.0 with an NVSwitched any-to-any network. For every environment, we repeat the same workload three times. 

\begin{figure}[h]
    \centering
    \begin{minipage}{0.3\textwidth}
        \includegraphics[width=0.96\textwidth]{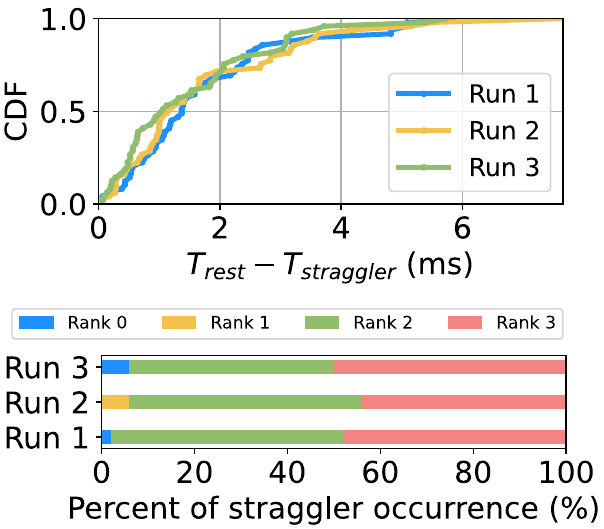}
        \caption*{(a) Perlmutter, 4 A100 SXM, Llama-3.2-1B, batch\_size=32.}
    \end{minipage}
    \hfill
    \begin{minipage}{0.3\textwidth}
        \includegraphics[width=0.98\textwidth]{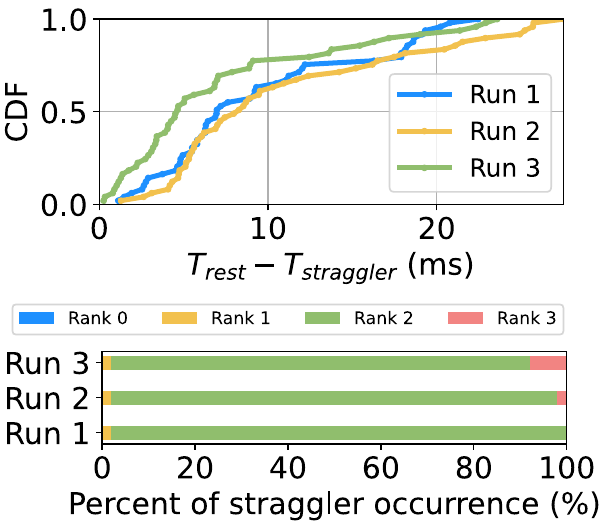}
        \caption*{(b) RunPod, 4 A100 SXM, Llama-3.2-3B, batch\_size=32.}
    \end{minipage}
    \hfill
    \begin{minipage}{0.3\textwidth}
        \includegraphics[width=\textwidth]{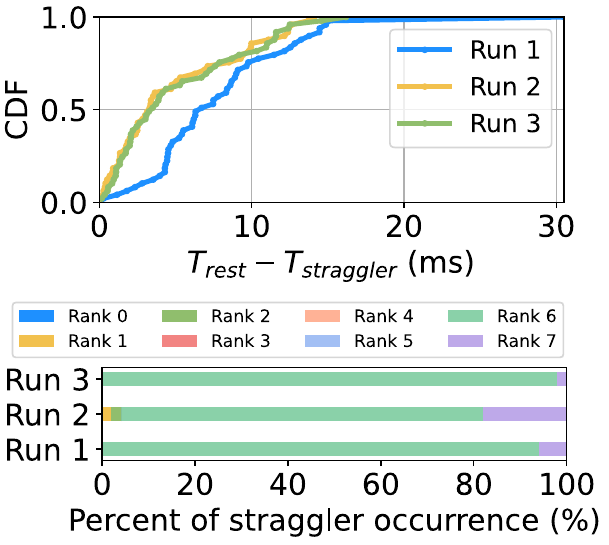}
        \caption*{(c) RunPod, 8 A100 SXM, Llama-3.2-3B, batch\_size=64.}
    \end{minipage}
    \caption{CDFs of the straggler delay, $T_{rest} - T_{straggler}$, and distribution of straggler ranks.}
    \label{fig:motivation}
\end{figure}

We encounter stragglers in all nine runs. We define the straggler delay as $T_{delay} = T_{rest} - T_{straggler}$, where $T_{straggler}$ is the \allreduce{} execution time of the straggler --- the GPU that started the \allreduce last (rank 0 in Fig.~\ref{fig:straggler-diagram}) --- and $T_{rest}$ is the \allreduce{} execution time of the second-slowest rank.
Note that $T_{straggler}$ is the shortest \allreduce{} time among all GPUs because the straggler spends no time waiting for others before communicating, and our measure of $T_{delay}$ is conservative since it only encodes the idle time of the second-slowest GPU.
We find a median straggler delay of more than 8 ms and a mean delay of 9.45 ms. The straggler GPU rank persists not only across iterations, but also across runs. Fig.~\ref{fig:motivation} shows that one specific GPU is the slowest in a communication group in 98\% of iterations, and is also the persistent straggler over multiple independent runs of the same workload on the same VM. Fig.~\ref{fig:motivation}(a) shows these findings also generalize to multiple stragglers.
Our experiments document (1) the severity of straggler GPUs that consistently delay \allreduce{}, and (2) that the same GPU rank is typically the straggler, allowing us to use offline profiling to bootstrap \sysname. These findings also suggest that routine stragglers may even appear from intrinsic and unavoidable GPU hardware heterogeneity (even with the same GPU model).

\section{Proof of Communication Complexity}
\label{subsec:proofs}
First, we prove that schedules synthesized by \sysname{} for power-of-2 $n$ are guaranteed to complete within 
$n {+}\log n{-}2$ rounds. As in Algorithm~\ref{alg:stragglar-appendix}, $A$ represents 
the data structure that maps each active chunk to the set of non-straggler ranks that holds it (and is updated in each round).
We assume this data structure is a hash table for constant-time updates and access. Note that an \emph{active chunk}
necessarily implies that the chunk has been fully reduced. Our code generates results that match the proof bounds.

\myparab{Lemma 1.} \emph{Prior to round $\log n$, every non-straggler rank possesses exactly one active chunk such that for $j = 0,\ldots,\log n {-}1, |A[c_{j}]| = 2 ^ {\log n - 1 - j}$.}
\label{lemma:base}

\begin{proof}
    After round $j < \log n$ has completed, $|A[c_{j}]| = 1$ because $c_{j}$ has just been fully reduced (via a pairing between rank $j$ and the straggler), and only rank $j$ holds it. 
    In round $j+1$, rank $j$ sends $c_j$ to rank $j + \log n$, so $|A[c_j]| = 2$.
    In each subsequent round $r = j + i$ for $i < \log n - j$, all ranks in $A[c_{j}]$ send $c_{j}$ to distinct ranks lacking a chunk, thus doubling $|A[c_{j}]|$ before the next round $r+1$. 

    Therefore, the number of non-straggler ranks possessing $c_j$ after round $r$ for $j \leq r < \log n$ is $|A[c_{j}]| = 2^{r - j}$.  After any round $r < \log{n}$, 
    the total number of non-straggler ranks holding an active chunk is 
    \[
    \sum_{j=0}^{r} 2^{r - j}
    \]
    We show that there are enough non-stragglers without a chunk after each round $r < \log{n} - 2$ (equivalently, before each round $r < \log{n} - 1$) to ensure that every rank with an active chunk has a distinct target it can send to:
    \[
        \sum_{j=0}^{r} 2^{r - j} \leq (n-2) - \sum_{j=0}^{r} 2^{r - j}
    \]

    ($n{-}2$ appears due to the straggler pairing in round $r$). Substituting the closed form of the sum,
\[
2^{\,r+1}-1
   \leq (n-2)-(2^{\,r+1}-1)
   = n-2^{\,r+1}-1,
\]
which simplifies to
\[
2^{\,r+2} \leq n.
\]
This inequality is only violated when $r{+}2 > \log n$. Thus, after each round $r \leq \log n{-}2$ (before each round $r \leq \log n{-}1$) there are strictly more
chunk-free non-stragglers than senders, so every sender can choose a distinct receiver.
    After round $r{=}\log n{-}1$ (\ie{}
    last round of Phase 1), $|A[c_{j}]| = 2 ^ {\log n - 1 - j}$, so the total number of non-straggler ranks with an active chunk is given by
    \[
    \sum_{j = 0}^{\log n - 1} 2 ^{\log n - 1 - j} = \sum_{i = 0}^{\log n - 1} 2^{i} = 2^{\log n} - 1 = n - 1
    \]
    This means all $n{-}1$ non-straggler ranks have a single fully reduced chunk after Phase 1.
\end{proof}

\myparab{Lemma 2.} \emph{Each chunk $c_r, r < n{-}1$, is fully reduced in round $r$, and is fully propagated to all ranks immediately after round $r + \log n$.}
\label{lemma:inductive}

\begin{proof}
    The first part of this statement follows directly from the schedule, as it dictates that partially reduced chunk $c_r$ resides on rank $r$ given
    the \reducescatter{} precondition, and rank $r$ communicates with the straggler, $\sigma$, in round $r$, enabling $c_r$ to be fully reduced 
    on ranks $r$ and $\sigma$. 

    To prove the second part of this statement, we first inductively prove the following invariant.

    \medskip
\noindent
Invariant $\mathcal{I}(r)$ holds immediately \emph{prior} to round $r \geq \log n$:
\begin{enumerate}[label=(\arabic*)]
  \item $|A[c_{j}]| = 2 ^ {r - j - 1}, \quad j = r{-}\log n,\ldots,r{-}1$
  \item $P_r = A[c_{\,r-\log n}]$ with $|P_r|=\tfrac{n}{2}$ and $r\in P_r$;
  \item $Q_r = \displaystyle\bigcup_{j=1}^{\log n-1} A[c_{\,r-j}]$ with $|Q_r|=\tfrac{n}{2}-1$;
  \item Active chunks are pairwise disjoint: $A[c_i] \cap A[c_j] = \varnothing, i\neq j,\; \forall i,j \in \{\,r{-}\log{n},{\dots},r{-}1\}$.
\end{enumerate}

    This enables us to match $u \in P_r \setminus \{r\}$ with $v \in Q_r$ to ensure that chunk $c_{r}$ has been 
    fully propagated to all ranks immediately after round $r + \log n$.
    
    \textbf{Base case.} We rely on Lemma~\hyperref[lemma:base]{1} for the base case. Immediately before round \(\log n\) (after round \(\log n{-}1\)), 
    \(|A[c_j]| = 2^{\log n - 1 - j}\), where sets \(A[c_j]\) are pairwise disjoint for all \(j = 0, 1, \dots,\log{n}{-}1\).
    Thus, $|P_{\log n}| = |A_{c_0}| = 2^{\log n - 1} = \tfrac{n}{2}$ and $|Q_{\log n}| = n{-}1{-}\tfrac{n}{2} = \tfrac{n}{2}{-}1$. Since rank $\log n$ 
    holds chunk $c_0$, as the protocol in Algorithm~\ref{alg:stragglar-appendix} dictates that rank $0$ sends $c_0$ to rank $\log n$ in round $1$, we 
    know $\log n \in P_{\log n}$. By matching $u \in P_{\log n}{\setminus}\{\log n\}$ with $v \in Q_{\log n}$ and having $u$ and $v$ each send their active chunks, we ensure that $c_{0}$ has 
    fully propagated after round $r={\log n}$. A matching is possible because every rank in $P_{\log n}$ lacks the active chunk of every 
    rank in $Q_{\log n}$, and vice versa. After the round, $c_{0}$ has fully propagated and every other active chunk has doubled.

    \textbf{Inductive step.} Assume $\mathcal{I}(r)$ holds immediately before round $r$ and that in round $r$
    we match some $u\in P_r\!\setminus\!\{r\}$ with a distinct $v\in Q_r$.
    Because $u$ sends $c_{r-\log n}$ to $v$, we have:
    \begin{enumerate}
      \item \emph{Full propagation.}\;
            After round $r$, every non-straggler now holds
            $c_{r-\log n}$ (since all ranks in $P_{r} \bigcup Q_r$ now have it), so this chunk is no longer active.
      \item \emph{Doubling of the remaining active chunks.}\;
            Every other active chunk is sent and received exactly once,
            so its cardinality doubles; every rank still stores exactly one
            active chunk. If a rank possessed the active chunk that just expired, it 
            received a new one, allowing it to maintain having exactly one active chunk. 
            If a rank possessed another active chunk, it received the active chunk that 
            just expired, allowing it to maintain the same status.
    \end{enumerate}

    We verify the validity of $\mathcal{I}(r{+}1)$ before round $r{+}1$ as follows. We can remove $c_{r - \log n}$ from 
    $A$ because it has fully propagated via the matching between $P_r$ and $Q_r$ in round $r$. We add $c_r$ to $A$ such that 
    $A[c_r] = \{r\}$, since it has just been fully reduced in round $r$ via the straggler pairing. Every other active chunk has 
    doubled in propagation (\ie the number of ranks that possess it) via the perfect matching between $P_r$ and $Q_r$. Let $j' = r{+}1{-}\log n$.
    Immediately \emph{before} round $r{+}1$ we verify:
    
    \begin{enumerate}[label=(\arabic*)]
      \item For every $k\in\{j',\dots,r{-}1\}, |A[c_k]| = 2|A[c_k]|_{\text{(before round }r)} = 2(2^{r-k-1}) = 2^{r-k}$
      \item By definition $P_{r+1}=A[c_{j'}]$. Via (1), we have $|P_{r+1}| = 2^{\log n-1}=\tfrac{n}{2}$. We prove that $r{+}1 \in P_{r+1}$ separately below.
      \item The cardinality of the remaining $\log n-1$ active chunks ensures the $Q_{r+1}$ constraint:
            \[
             |Q_{r+1}| = \sum_{k=j'+1}^{r} |A[c_k]| = \sum_{t=1}^{\log n-1} 2^{\log n-1-t} = \frac{n}{2}-1
            \]
      \item Since each rank still holds exactly one active chunk,
            the sets $A[c_k]$ remain pairwise disjoint. In round $r$, ranks in $Q_r$ received $c_{r - \log n}$, which is no longer active by round $r{+}1$, so 
            these ranks still retain the same active chunk they had prior to round $r$. Ranks in $P_r \setminus \{r\}$ had only active chunk $c_{r - \log n}$ before round $r$, which 
            expired as an active chunk, and received some other active chunk in round $r$, so they also only have one active chunk each. Finally, $r \in P_r$, so its active chunk 
            expired after round $r$, and it received active chunk $c_r$ via the straggler pairing, so it also only has one active chunk. This ensures pairwise disjointness.
    \end{enumerate}
    
    There are two remaining components of the proof for the inductive step: (1) that $r{+}1 \in P_{r+1}$, and (2) that there always exists a feasible matching between $P_{r+1} \setminus \{r+1\}$ and 
    $Q_{r+1}$. First, it is guaranteed that $r{+}1 \in P_{r+1}$ because of the rule for matching ranks in the critical window from Algorithm~\ref{alg:stragglar-appendix}. Specifically, we ensure that rank $r{+}1$ is never paired with a rank whose 
    active chunk is $c_{l}$ for $l > r{+}1{-}\log n$. This rule ensures that $r{+}1 \in P_{r+1}$ because the only active chunk it could have at the beginning of round $r{+}1$ is then $c_{r+1-\log n}$, guaranteeing 
    that $r{+}1 \in P_{r+1}$. 

    The next question is whether we can ensure that there exists a matching in round $r{+}1$. First, we need to ensure that any rank $l$ in the critical window is not matched with a 
    rank whose active chunk will continue to be active by round $l$. If $l \in Q_{r+1}$, this is trivial because the rank does not receive any new active chunk in that round (since it only 
    receives the chunk that is due in that round). If $l \in P_{r+1}$, we just have to ensure that $l$ is not matched with any of the ``newest'' active chunks; in fact, matching $l$ with 
    ranks possessing the second-oldest active chunk always works to keep $l$ in future $P_{r}$. Since $G_r$, the graph 
    that connects $P_r$ and $Q_r$, is a complete bipartite graph, we can easily guarantee this pairing for any $l$. By removing these $l$ and their partners from their respective sets (either $P_r$ or $Q_r$), 
    we can trivially apply Hall's Marriage Theorem since $|P_r| = |Q_r|$ (and $G_r$ is bipartite) to verify that a matching exists. Hence, $\mathcal{I}(r{+}1)$ holds, completing the inductive step.
\end{proof}

\myparab{Remark 1} \emph{The final chunk, $c_{n-2}$, propagates in $\log n - 1$ rounds.}
\label{remark}

After round $n{-}2$, the straggler, $\sigma$, now has all fully reduced chunks, and is no longer constrained by whom to pair with. This means 
we can add the straggler to $Q_r$ for subsequent rounds $r \geq n{-}1$ such that $|P_r|=|Q_r|=\tfrac{n}{2}$. (We enforce the invariant for subsequent rounds that the straggler can only send $c_{n-2}$.) Both ranks $n{-}2$ and $\sigma$ are in $Q_r$ (and will remain in $Q_r$ until the end 
of the algorithm), and their active chunk is $c_{n-2}$. Thus, $c_{n-2}$ begins with 2 ranks holding it, as opposed to 1, jumpstarting its doubling process. Every 
chunk can continue to double because $|P_r|=|Q_r|$ and every rank only has one active chunk, so via Hall's Marriage Theorem a perfect matching exists.
Therefore, only $\log n - 1$ rounds are required to fully propagate the final chunk.\\

\myparab{Theorem 1.} \emph{\allreduce schedules generated by \sysname complete in $n{+}\log n{-}2$ rounds.}
\begin{proof}
    Naturally, $n{-}1$ rounds are required for all chunks to become fully reduced, via the linear sequence of pairing with the straggler, 
    with chunk $c_{j}$ fully reduced in round $j$.
    Lemma~\hyperref[lemma:inductive]{2} proves that each chunk $c_{j}$ propagates by round $j{+}\log n$. Thus, the total number of rounds in the 
    algorithm is bounded by the final chunks. Chunk $c_{n-3}$ will propagate by round $n{-}3{+}\log n$ and chunk $c_{n-2}$ will have fully propagated after 
    round $n{-}2{+}\log n{-}1 = n{-}3{+}\log n$. Since we zero-index rounds, this results in $n{+}\log n{-}2$ rounds in total.
\end{proof}

\section{\sysname{} for Non-Power-of-$2$ Values of $n$}
\label{sec:nonpow2}
When the number of GPUs is not a power of $2$, \sysname{}
becomes more challenging to implement because GPUs can possess multiple active chunks at any point in time.
To handle the even, non-power-of-$2$ case, we use a similar matching-based approach to construct the 
schedule in every round, but instead connect vertices $u$ and $v$ with weight $2$ if $u$ possesses 
a chunk that $v$ needs and $v$ possesses a chunk that $u$ needs, and with weight $1$ if only $u$ 
possesses a chunk that $v$ needs, but the opposite is not true (or vice versa). Then, we run well-known
polynomial-time algorithms~\citep{edmonds1965paths, edmonds1965maximum} to compute the maximum weight matchings per round in order to generate the schedule.
Thus, offline schedule generation when $n$ is not a power-of-2 has a higher runtime, but is still polytime computable.

\begin{figure}[h]
    \centering
    \includegraphics[width=0.6\textwidth]{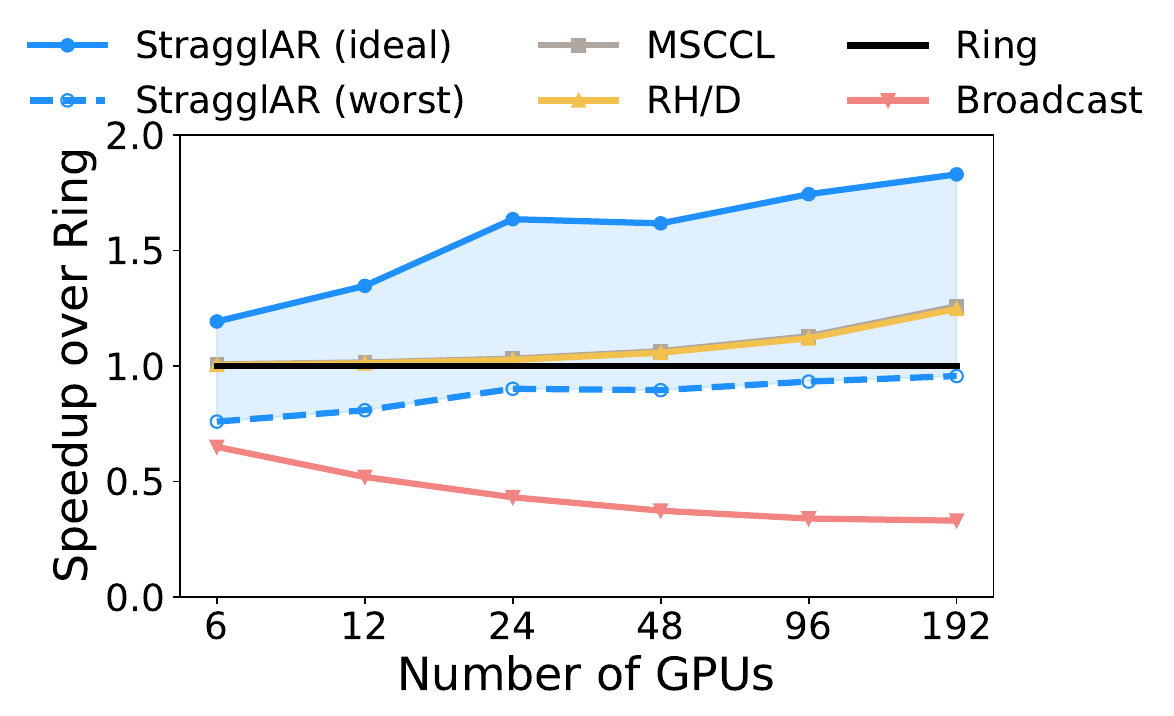}
    \caption{\allreduce performance of different algorithms as cluster size $n$ scales, for even values of 
    $n$ that are not powers of 2. The shaded region captures the range of \sysname{'s} possible performance, ranging from worst case (no straggler and hence no overlap) to ideal (full overlap with straggler).}
    \label{fig:nonpow2}
\end{figure}

For even, non-power-of-2 values of $n$, we do not prove a bound on the $\alpha{-}\beta$ cost. However, we still synthesize and 
evaluate these schedules in \S\ref{sec:eval} and typically find schedules that complete in 
${\sim}n{+}2\log n{-}2$ rounds in practice, which still results in lower $\beta$ cost than baselines. The realized communication complexity of the
schedules generated by \sysname{} for non-power-of-2 $n$ is shown in the ideal-case performance of Fig.~\ref{fig:nonpow2}, and still outperforms all baselines
using the analytical network simulator. We note that \sysname{} does not provide schedules for odd values of $n$, but these settings are less
common in large-scale distributed ML; further, \sysname{'s} schedules for even $n$ are all computable in polytime due to the graph matching structure.

We compare \sysname{'s} ideal and worst-case performance to baselines as the cluster size scales, in simulation with the analytical network model. 
We report the results in Fig.~\ref{fig:nonpow2}, which shows a similar pattern as for powers-of-2. In ideal scenarios, when the \reducescatter{} 
precondition can be fully overlapped with the straggler delay, \sysname{} significantly outperforms all baselines. For worst-case scenarios with 
no straggler delay, \sysname{'s} performance is slightly worse than baselines at small cluster sizes, but performs equally well at large cluster sizes 
due to its $2\times$ lower asymptotic communication complexity.

\section{Discussion of Baselines}
\label{sec:baselines}
We provide in-depth explanations and $\alpha{-}\beta$ costs of the baselines. Classical algorithms (\eg Ring, RH/D) 
achieve the known lower bounds for \allreduce in networks with homogeneous bandwidth and thus remain the standard implementations 
in state-of-the-art CCLs~\citep{hu2025demystifying}. While newer works synthesize algorithms for irregular topologies and \emph{heterogeneous networks}~\citep{won2024tacos,wang2020blink,zhao2024adapcc},
they do not provide new algorithms for homogeneous scale-up networks, aside from MSCCL~\citep{cowan2023mscclang}, which we compare to.

 \myparab{Ring~\citep{patarasuk2009bandwidth}.} This is the standard bandwidth-efficient algorithm implemented by NCCL for \allreduce{}. 
 It divides the buffer into chunks of size $\tfrac{s}{n}$ and consists of $2(n-1)$
 rounds in a ring-like communication pattern. The total runtime is $T_{Ring} = 2(n-1)\cdot \alpha + 2\tfrac{n-1}{n}s\cdot\beta$.

 \myparab{Recursive halving/doubling [RHD]~\citep{bruck1994efficient}.} Also called the \emph{Butterfly} algorithm, 
 this algorithm---like Ring---consists of two phases. In the first phase, the buffer is first divided in $1/2$ with one partner, 
 then in $1/4$ with another partner, \etc{} to achieve a \reducescatter{}. Then, a mirror-image \allgather{} completes to propagate 
 all fully reduced chunks. The total runtime is $T_{RHD} = 2 \log n \cdot \alpha{+}2\tfrac{n-1}{n}s \cdot \beta$.

\myparab{Broadcast.} This is a \emph{straggler-aware}
 baseline that assumes the non-stragglers complete 
 an \allreduce{} ($2\times$ the time as \reducescatter{}) during the straggler delay. 
 After the straggler is ready, it exchanges its \emph{entire} buffer with any other rank to fully reduce the entire buffer and then initiates a broadcast with $\log n$ rounds and $s$ bytes per round. The total ideal-case runtime is $T_{Bcast} = \log n \cdot \alpha + \log n \cdot s \cdot \beta$
 while the worst-case runtime is $(2n + \log n - 4) \cdot \alpha + (\log n + 2\frac{n-2}{n-1}) \cdot s \cdot \beta$.

\myparab{MSCCL~\citep{cowan2023mscclang}.} This is a new algorithm synthesized for switched scale-up domains, such as NVIDIA DGX. The algorithm, called \emph{allpairs}, consists of just two rounds. The buffer is divided into $n$ chunks.
In the first round, each GPU is responsible for reducing one chunk (\ie \reducescatter), and sends all other chunks to other GPUs simultaneously, thereby splitting the link bandwidth among $n{-}1$ parallel connections. The next step is a 
mirror-image \allgather, where all GPUs communicate simultaneously to receive all reduced chunks. Thus, the total runtime is $T_{MSCCL} = 2 \cdot \alpha{+}2\tfrac{n-1}{n}s \cdot \beta$. 

\section{Evaluation on Perlmutter with 4 GPUs}
\label{sec:perlmutter}
\begin{figure}[h]
    \centering
    \includegraphics[width=\linewidth]{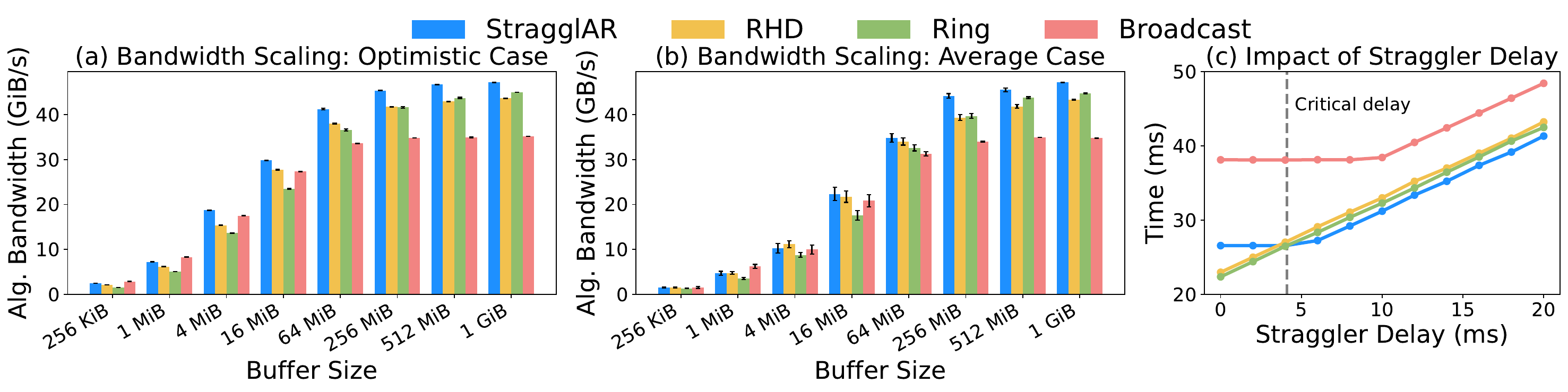}
    \caption{\allreduce{} performance on a 4-GPU node in Perlmutter. (a) shows the optimistic use case, where  
     it is assumed that the \reducescatter{} can complete within the straggler delay time. (b) assumes a 
     straggler delay of 9.45 ms, the average from our experiments. (c) fixes the buffer size at 1 GiB and varies the straggler delay.}
    \label{fig:4gpu}
\end{figure}
We show the results from running on one node of the Perlmutter supercomputer~\citep{nersc_perlmutter_architecture} in Fig.~\ref{fig:4gpu}.
While \sysname{} still consistently outperforms baselines across buffer sizes in the 4-GPU setup, its gains are less substantial than for the 8-GPU setup. 
This is expected, and the results are in line with the theoretical results in \S4 because \sysname{'s} performance advantage \emph{improves} as 
$n$, the size of the GPU cluster, scales. (Intuitively, this is because the proportion of  ``work'' that is offloaded to the precondition is higher as $n$ scales.)
Fig.~\ref{fig:4gpu}b indicates similar performance on the 4-GPU node even with both the average and masked delays, suggesting 
the \reducescatter{} precondition can be effectively masked in typical scenarios.

\section{NCCL Tuning Behavior}
\label{subsec:nccl-tests}
\begin{figure}[h]
    \centering
    \includegraphics[width=0.42\linewidth]{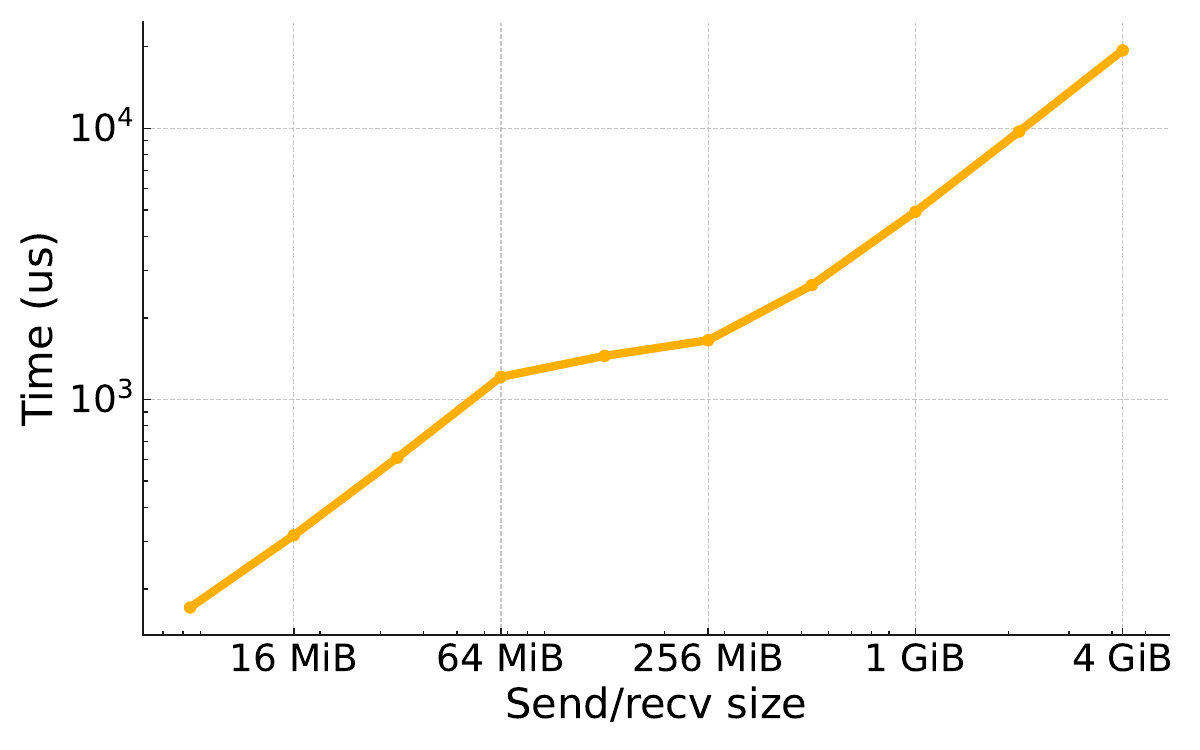}
    \caption{\texttt{nccl-tests} benchmarking of NCCL P2P API on RunPod server with 8 A100 GPUs.}
    \label{fig:nccl-tests}
\end{figure}

As we discuss in \S\ref{sec:eval}, the performance of \texttt{ncclSend()}/\texttt{ncclRecv()} 
deviates from the $\alpha{-}\beta$ cost model's expected performance for some specific buffer sizes 
due to heuristics and protocol-switching behavior implemented internally by NCCL. We run the 
baseline \texttt{nccl-tests}~\citep{nccl_tests}, a benchmarking suite provided by NVIDIA for evaluating in-built NCCL performance, across different data sizes and plot the results in Fig.~\ref{fig:nccl-tests}. The region from 64 MiB-256 MiB deviates from 
expected behavior under $\alpha{-}\beta$ costs because sending $2\times$, and even $4\times$, the amount of data results in almost no change in 
communication time with the P2P API. (Under the $\alpha{-}\beta$ model, the time should consistently scale linearly with the send/recv data size, instead of step-wise as shown in Fig.~\ref{fig:nccl-tests}.) 
This is why the Direct and MSCCL \allreduce baselines outperform other algorithms at 256 MiB, as the time to send an entire 256 MiB buffer 
is only marginally higher than sending smaller chunk sizes, which bandwidth-efficient algorithms like Ring, RHD, and \sysname do. 
We do not observe this finding to generalize to other architectures, and it may be a relic of the NCCL version, CUDA version, and specific 
environment configured by RunPod for \texttt{ncclSend()}/\texttt{ncclRecv()}. Also, outside of the 64-256 MiB range, the performance closely reflects 
the $\alpha{-}\beta$ model, and our findings in \S\ref{sec:eval} reflect this. This anomalous behavior is also well documented by recent works, which even propose methods to optimize tuning parameters for NCCL~\citep{hu2025demystifying,wang2025simai}.

\section{End-to-End Training Experiments on Multi-GPU Servers}
\label{sec:e2e-evals}
We evaluate \sysname{} using data-parallel (DP) fine-tuning of the Llama3.2-3B model in PyTorch on a DGX A100 VM (8-GPU setup) on RunPod. In this evaluation, we enable \allreduce algorithms to 
run in the wild, calling implementations directly from PyTorch on a real hardware testbed. To do so, we compile our C++ and 
CUDA implementations (using the NCCL P2P API, exactly as described in \S\ref{sec:eval}) into a custom package, 
and enable \texttt{dist.all\_reduce()} to call these implementations using PyTorch Distributed's C++ extension capabilities~\citep{pytorch_cpp_extension}. 
We compare \sysname{} to the Ring algorithm, the most well-known and widely used bandwidth-efficient baseline, 
under this scheme. Using standard PyTorch profiling tools, we identify device (GPU) 6 to 
be the most frequent straggler on our server, as device 6 was most frequently the straggler in the profiling logs. At the start of training, we pass the straggler rank to the CUDA runtime to ensure \sysname{} executes correctly. 
Then, training simply proceeds as per usual with PyTorch.

\begin{figure}[H]
    \centering
        \includegraphics[width=0.45\textwidth]{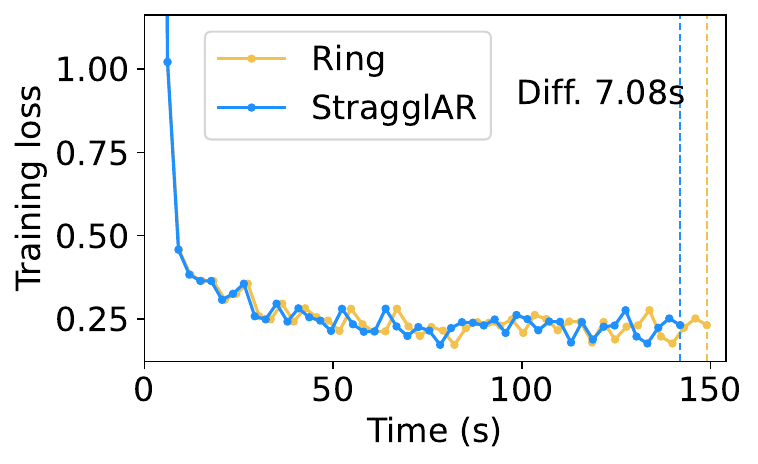}
    \caption{End-to-end evaluation for Llama-3.2-3B fine-tuning on a RunPod DGX A100.}
    \label{fig:ar-loss}
\end{figure}

Next, we fine-tune Llama3.2-3B using the same data-parallel setup on 8 GPUs with the global batch size of 32 
for 100 iterations to determine the overall training efficiency gains with \sysname{}. Fig.~\ref{fig:ar-loss} shows that 
even these first 100 iterations, \sysname{} reduces training time by 7 seconds. These gains suggest that training 
at longer time scales---for example, days---can shave off several hours of training.
Further, we note that \sysname{'s} loss curve is just the Ring \allreduce{} curve shifted 
to the left: the $y$-coordinates are exactly the same because both algorithms ensure \emph{complete} data buffers are reduced, ensuring  
no change in model convergence. Instead, \sysname{} enables an unmodified, but faster \allreduce{} to the application layer, resulting in greater training efficiency and faster time-to-convergence. Fig.~\ref{fig:ar-loss} captures 
the wall-clock time, \ie{} all compute and communication operations in training, so the results may improve in larger-scale settings 
(as shown in simulations in \S\ref{sec:e2e-simulation}) or with even larger model sizes where communication becomes a greater bottleneck.

\section{End-to-End Training Simulations at Scale}
\label{sec:e2e-simulation}
\begin{figure}[h]
    \centering
    \includegraphics[width=0.9\linewidth]{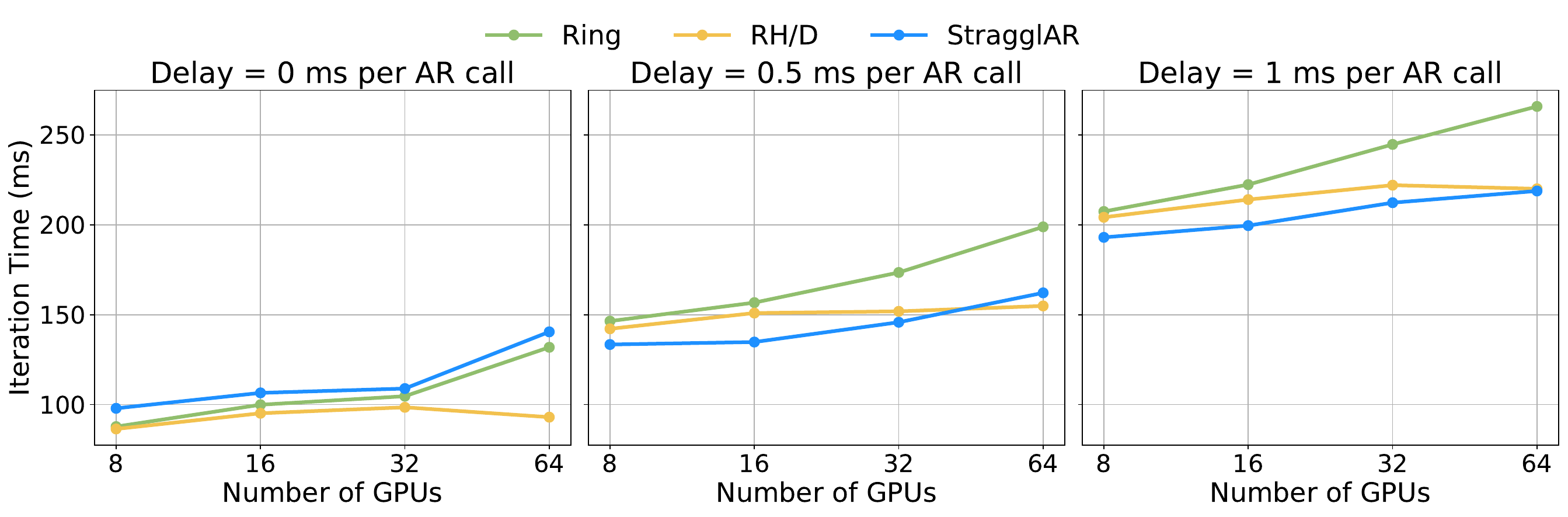}
    \caption{Runtime per BERT training iteration using the FlexNet simulator with $\alpha{=}3 \mu{s}$.}
    \label{fig:e2e-sim}
\end{figure}

\begin{figure}[H]
    \centering
    \includegraphics[width=0.9\linewidth]{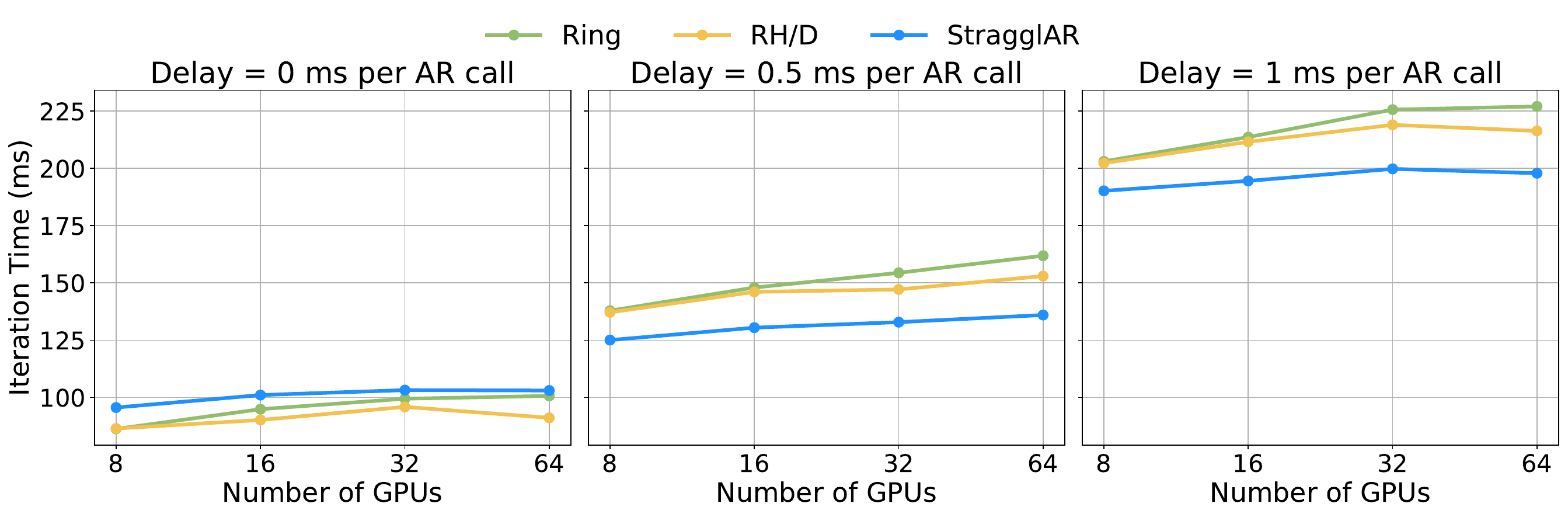}
    \caption{Runtime per BERT training iteration using the FlexNet simulator with $\alpha{=}0.7 \mu{s}$.}
    \label{fig:e2e-sim-taccl}
\end{figure}

We use the FlexNet simulator used by prior work~\citep{wang2023topoopt}---an augmented version of 
FlexFlow~\citep{jia2019beyond, flexflow}---to determine 3D parallelism schemes for transformer training at even larger scales.
Specifically, FlexNet outputs the task graph of compute and communication tasks per training iteration.
We simulate training the BERT-large transformer model~\citep{kenton2019bert} as we vary both the number of GPUs and the 
straggler delay. We provide end-to-end training iteration time reported by FlexNet and using 
the analytical network simulator to predict the communication time (for both collective and point-to-point communication) as an input.
Fig.~\ref{fig:e2e-sim} shows end-to-end results assuming $\alpha{=}3 \mu{s}$ (based on ~\citep{nvlink_latency}) while Fig.~\ref{fig:e2e-sim-taccl}
shows these results assuming $\alpha{=}0.7 \mu{s}$ (based on ~\citep{taccl}). In general, 
\sysname{} outperforms baselines as the number of GPUs scales when there is even marginal delay of 0.5 ms or higher.
When there is no delay, \sysname suffers because \emph{none} of the \reducescatter precondition is overlapped 
with the straggler delay. In Fig.~\ref{fig:e2e-sim}, \sysname either performs more poorly than or evenly with RHD (depending on the straggler delay) 
for $n{=}64$ because the assumption of $\alpha$ is higher (so, algorithms like RHD that minimize the 
number of rounds are preferred when $n$ is large). In contrast, Fig.~\ref{fig:e2e-sim-taccl} assumes a 
lower value of $\alpha$ that is reported by prior published work~\citep{taccl}, thereby enabling \sysname{'s} bandwidth efficiency to shine
even at large values of $n$. The exact buffer sizes and compute and communication kernels are directly outputted by 
FlexNet, and we evaluated with batch sizes of 16, 32, 64, and 128 on 8, 16, 32, and 64 GPUs, respectively. These 
batch sizes are based on typical fine-tuning use cases that opt for smaller batch sizes on smaller datasets~\citep{chung2024scaling, liu2019roberta,touvron2023llama}.
\end{document}